\documentclass[lettersize,journal]{IEEEtran}
\usepackage{amsmath,amsfonts}
\usepackage{array}
\usepackage{textcomp}
\usepackage{stfloats}
\usepackage{verbatim}

\usepackage{threeparttable}
\usepackage{cite}
\usepackage{ulem}
\usepackage{caption}
\usepackage{graphicx}
\usepackage{multirow}
\usepackage{subfigure}
\usepackage{lineno}
\usepackage{times}
\usepackage{helvet}
\usepackage{courier}
\usepackage{epstopdf}
\usepackage{color}

\usepackage{ntheorem}
\usepackage{bm}
\usepackage{graphicx}
\usepackage{subfigure}
\usepackage{float}  
\usepackage[linesnumbered,lined,ruled]{algorithm2e}
\usepackage{adjustbox}

\theoremheaderfont{\bfseries \itshape}
\theorembodyfont{\upshape}
\newtheorem{assumption}{Assumption}
\newtheorem{definition}{Definition} 
\newtheorem{theorem}{Theorem}
\newtheorem{lemma}{Lemma}[section]

\newtheorem{game}{Game}

\newenvironment{proof}{%
\vspace{-0.2cm}
\textit{Proof.} 
}{\hfill $\square$ }

\makeatletter

\newcommand{\Rmnum}[1]{\expandafter\@slowromancap\romannumeral #1@}
\makeatother

\DeclareMathOperator*{\argmax}{arg\,max}
\DeclareMathOperator*{\argmin}{arg\,min}

\begin{document}

\captionsetup{font={footnotesize}}

\title{Multi-Time Scale Service Caching and Pricing  in MEC Systems with Dynamic Program Popularity}

\author{ 
		Yiming~Chen,
		Xingyuan~Hu,
		Bo~Gu,~\IEEEmembership{Member,~IEEE},
		Shimin~Gong,~\IEEEmembership{Member,~IEEE},
		Zhou~Su,~\IEEEmembership{Senior Member,~IEEE}
	
	\thanks{This work was supported in part by the National Science Foundation of China (NFSC) under Grant U20A20175. (\textit{Corresponding author: Bo Gu.})}
	
	\IEEEcompsocitemizethanks{
		\IEEEcompsocthanksitem Yiming Chen, Xingyuan Hu, Bo Gu and Shimin Gong are with the School of Intelligent Systems Engineering, Shenzhen Campus of Sun Yat-sen University, Shenzhen 518107, China (e-mails: chenym98@mail2.sysu.edu.cn; huxy86@mail2.sysu.edu.cn; gubo@mail.sysu.edu.cn; gong0012@e.ntu.edu.sg;).
		\IEEEcompsocthanksitem Zhou Su is with the School of Cyber Science and Engineering, Xi'an Jiaotong University, Xi'an 710049, China (e-mail: zhousu@ieee.org).
	}
}

\maketitle
\begin{abstract}
	In mobile edge computing systems, base stations (BSs) equipped with edge servers can provide computing services to users to reduce their task execution time. However, there is always a conflict of interest between the BS and users. The BS prices the service programs based on user demand to maximize its own profit, while the users determine their offloading strategies based on the prices to minimize their costs. Moreover, service programs need to be pre-cached to meet immediate computing needs. Due to the limited caching capacity and variations in service program popularity, the BS must dynamically select which service programs to cache. Since service caching and pricing have different needs for adjustment time granularities, we propose a two-time scale framework to jointly optimize service caching, pricing and task offloading. For the large time scale, we propose a game-nested deep reinforcement learning algorithm to dynamically adjust service caching according to the estimated popularity information. For the small time scale, by modeling the interaction between the BS and users as a two-stage game, we prove the existence of the equilibrium under incomplete information and then derive the optimal pricing and offloading strategies. Extensive simulations based on a real-world dataset demonstrate the efficiency of the proposed approach.
\end{abstract}
\section{Introduction}
\ULforem
\IEEEPARstart{W}{ith} the explosive development of computation-intensive applications (e.g., image recognition, virtual reality, and autonomous driving), the limited computation capability of devices often leads to significant delays in local computing, which decreases the quality of service (QoS) for users \cite{10261251}. Alternatively, mobile edge computing (MEC) is proposed as a promising paradigm to address the above issues \cite{sun2020bandwidth}. Users are able to offload their tasks to nearby base stations (BSs) equipped with edge servers that provide enhanced computing ability with low latency \cite{el2019joint}. According to the International Data Corporation, edge computing will have a large worldwide market in 2024, representing  an increase of 15.4\% by 2023 \cite{IDC2024}.

Despite these advantages, it is critical to address the following issues to achieve successful deployment of MEC systems. \textit{\textbf{Service Caching.}} The BS can provide computation service to users only if it pre-caches the corresponding service programs \cite{yan2021pricing}. If a program is not pre-cached, it costs tens of seconds to temporarily install it, which is unacceptably long compared to the task execution time \cite{zhao2018red}. Existing works \cite{yan2019optimal,tong2023stackelberg} assume that the BS caches all service programs, which is unrealistic due to the limited caching capacity. Therefore, the BS needs to selectively cache suitable programs. \textit{\textbf{Pricing.}} Without an effective incentive mechanism, self-interested users tend to occupy as many computing resources as possible. Therefore, it is essential for the BS to set prices
for different programs according to user demand. Appropriate prices can not only improve resource utilization but also increase profits. \textit{\textbf{Task Offloading.}} Self-interested users need to weigh the trade-off between time delay and payment expenditure to make optimal offloading decisions.

Jointly optimizing service caching, pricing, and task offloading  is a challenging task for the following reasons. First, service caching and pricing require different time granularities for updates. If pricing updates are excessively slow, it is impossible for the BS to set suitable prices to capture the rapid changes in user demand, potentially leading to a decrease in profits. Conversely, frequent adjustments in service caching may lead to excessive caching overhead  (including time consumption and price), and may even cause system instability \cite{thai2019workload}. Therefore, the BS must choose appropriate optimization time scales for service caching and pricing. Second, dynamic changes in program popularity require the popularity-aware ability of the BS, without this, the caching will have difficulty retrieving high-popularity programs. Furthermore, pricing and task offloading are coupled with each other; namely, the pricing of the BS must consider users' offloading behaviors, and the offloading strategies are directly related to the prices. Last but importantly, users typically do not disclose their private information to other users. Hence, each user must make offloading decisions under incomplete information. This further complicates the joint optimization problem.

To address the above issues, we present a two-time scale optimization framework for service caching, pricing, and task offloading in a MEC system. For the small time scale, we model the price competition between the BS and users as a Stackelberg game \cite{fudenberg1991game}, and derive the optimal offloading and pricing strategies. For the large time scale, a game nested deep reinforcement learning (GNDRL) algorithm is designed to determine the optimal service caching. The main contributions of this paper are summarized as follows:
\begin{itemize}
	\item \textbf{Two-time scale  optimization framework.} To balance the  needs for different adjustment time granularities, we propose a two-time scale optimization framework. Service caching is optimized on a large time scale while pricing and task offloading are optimized on a small time scale.
	\item \textbf{GNDRL algorithm for service caching.} For the large time scale, we formulate the service caching problem as a Markov decision process (MDP) and design the GNDRL algorithm  to determine the BS's optimal caching strategy according to the estimated program popularity.
	\item \textbf{Stackelberg game-based pricing and task offloading.} For the small time scale, we model the interaction between the BS adjustments users as a Stackelberg game. We prove the existence of the equilibrium and derive the users' optimal offloading strategies. Moreover, accounting for algorithm profitability and computation complexity, we design two pricing algorithms to obtain the optimal prices.

	\item \textbf{Extensive simulations on a real-world dataset.} We conduct numerical simulations to evaluate the performances (e.g., total cost of users, profit of the BS) of our proposed algorithms. Moreover, we employ real-world data to demonstrate the feasibility of our proposed algorithms in practical scenarios.
\end{itemize}

The remainder of the article is organized as follows. Related work is reviewed in Section \Rmnum{2}. In Section \Rmnum{3}, we introduce the system model. Section \Rmnum{4} formulates the optimization problems. Service caching based on GNDRL is introduced in Section \Rmnum{5}. In Section \Rmnum{6}, we analyze the Stackelberg game and derive the users' optimal offloading strategies and the BS's optimal pricing scheme. In Section \Rmnum{7}, the numerical simulations are described.  Section \Rmnum{8} concludes the paper.

\section{Related Work}

\subsection{Cooperative Task Offloading}
Existing work has extensively studied centralized optimization within MEC systems. In such scenarios, edge servers and users share a common optimization objective, typically including but not limited to, minimizing the computation latency or energy consumption of the entire MEC system. Wang \textit{et al.} \cite{wang2023delay} proposed a distributed algorithm using a mean field game to optimize the delay-optimal computation offloading problem. Yeganeh \textit{et al.} \cite{yeganeh2023novel} used an enhanced hybridization algorithm to optimize task offloading and scheduling. Fan \textit{et al.} \cite{fan2022joint}  proposed a resource management scheme and designed an iterative algorithm to optimize the delay and energy consumption of the system. Liu \textit{et al.} \cite{liu2023dependent} minimized the system deadline violation ratio to improve the overall reliability performance via task migration and merging. Dong \textit{et al.} \cite{dong2023joint} introduced non-orthogonal multiple access, studying power allocation and task offloading to reduce energy consumption for all devices. Although the aforementioned works conducted joint optimization of service caching, task offloading and resource allocation, they overlooked the fact that BSs and users, who are self-interested in real-world scenarios, prioritize their own benefits over overall system efficiency.

\begin{figure}[tbp]
	\vspace{-0.4cm}
	\centering
	\includegraphics[width=0.48\textwidth]{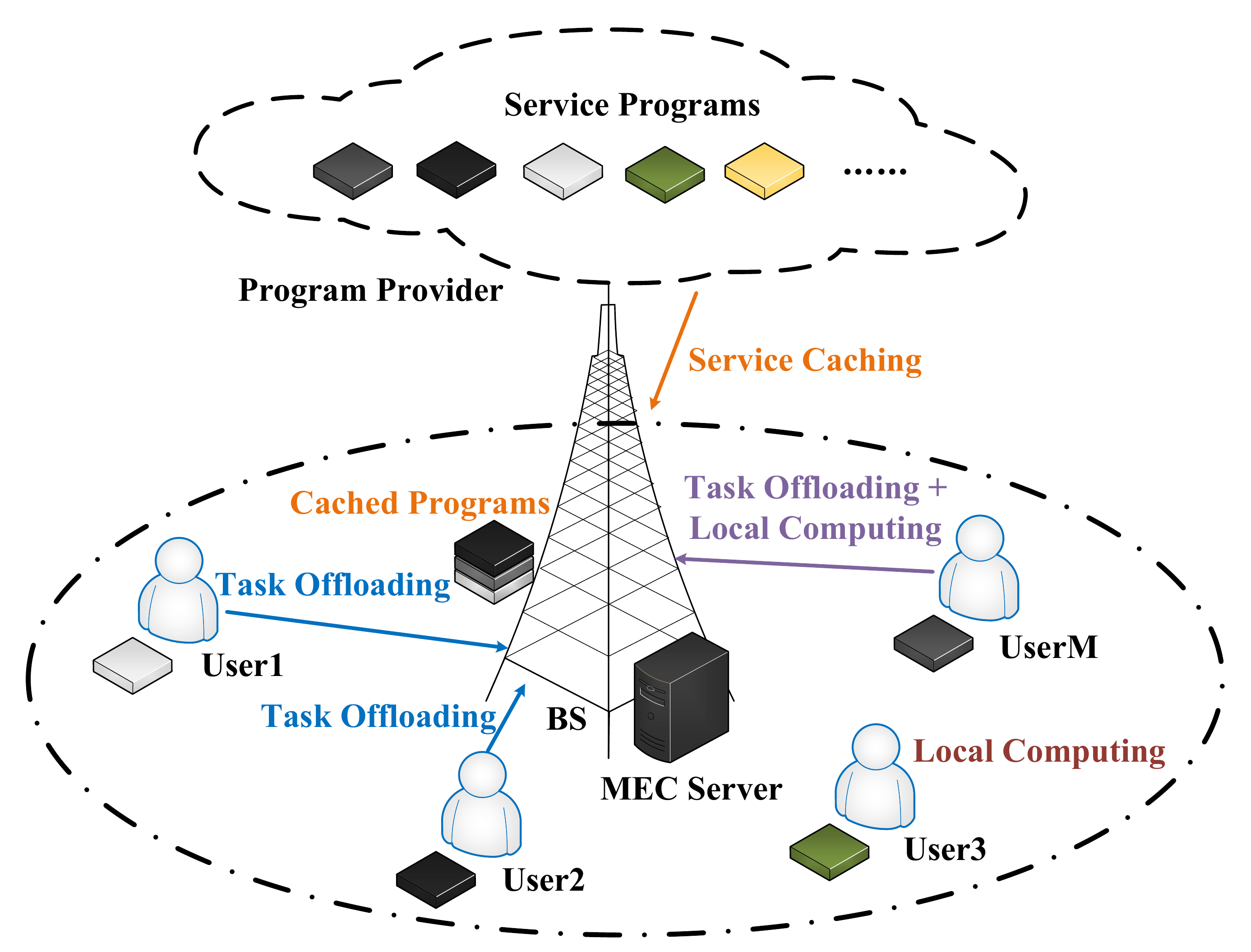}
	\caption{System model.}
	\label{fig:system}
	\vspace{-0.2cm}
\end{figure}

\subsection{Competitive Task Offloading}
To address the aforementioned issues, some studies have treated the BS and users as self-interested agents who optimize pricing and task offloading without considering the overall system latency or energy consumption, but rather aim to maximize their own benefits. Wang \textit{et al.} \cite{wang2019profit} introduced an online multi-round auction mechanism that maximizes profits through incentive mechanisms, satisfying user QoS while ensuring edge server profits. Liwang  \textit{et al.} \cite{liwang2022unifying} created a hybrid market between the BS and users, aiming to reduce trading decision-making delays and enhance resource utilization. Tong \textit{et al.} \cite{tong2024stackelberg} employed a Stackelberg game to analyze the interaction process between an edge server and users, and established an optimal relationship between bandwidth and task size to optimize bandwidth allocation and resource pricing. The above studies considered single-edge service MEC scenarios, whereas Chen \textit{et al.} \cite{chen2022price} investigated a heterogeneous edge servers model, considering price competition among multiple servers on top of the server-user game. In addition, Huang \textit{et al.} \cite{huang2023pricing} optimized the server configuration and overload in a multi-server scenario, in which BSs dynamically switch servers on/off and purchase idle resources from private users to maximize resource utilization. The above works studied the decision-making behavior of edge servers and users under competitive relationships, but neglected the pre-caching of service programs by the BS during the computation offloading process.

\subsection{Service Caching}
Some studies have included service caching optimization into MEC system analysis. Chu \textit{et al.} \cite{chu2023joint} studied the problem of maximizing users' QoS, by jointly optimizing service caching, resource allocation and task offloading. Qin \textit{et al.} \cite{qin2023joint} proposed a user-centric edge service caching framework and calculated the long-term average delay under the constraint of the caching cost minimization problem. Xu \textit{et al.} \cite{xu2022stable} exploited stable service caching via a Stackelberg game in a request rate uncertainty scenario. These works only considered static MEC caching scenarios, whereas to adapt to the system's dynamic nature, a multitude of studies have introduced algorithms for dynamically updating caching strategies based on partial observation information. For instance, Farhadi \textit{et al.} \cite{farhadi2021service} jointly optimized service placement and request scheduling under a two-time scale framework. Ren \textit{et al.} \cite{ren2022adaptive} also utilized the two-time scale framework, but considered a multi-server scenario to optimize request scheduling and service caching. Yao \textit{et al.} \cite{yao2023cooperative} constructed a digital twin edge network to reflect dynamic system features, and proposed a graph attention network-based reinforcement learning algorithm to learn optimal task offloading and service caching strategies.

The aforementioned works have contributed significantly to the research on static and dynamic service caching in MEC systems, but have engaged less with  the dynamic awareness of service program popularity. Moreover, few studies have effectively combined service caching with competitive games between edge servers and users. Therefore, this paper aims to investigate service caching and pricing in a MEC system based on the dynamic nature of program popularity and user demand while optimizing the offloading strategies of users.

\section{System Model}
\begin{figure}[tbp]
	\centering
	\includegraphics[width=0.42\textwidth]{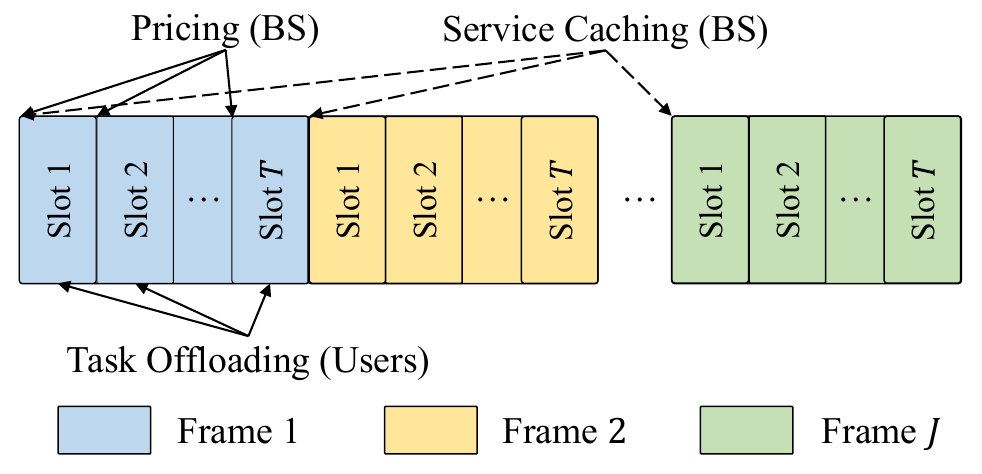}
	\caption{Two-time scale framework for service caching, pricing and task offloading.}
	\label{fig:tts}
	\vspace{-0.2cm}
\end{figure}

As shown in Fig. \ref{fig:system}, we consider a  MEC system with single BS and $M$ users. The set of users is denoted by $\mathcal{M}=\left\{1,\cdots,M\right\}$. The BS is equipped with an edge server that provides computing resources for users. As shown in Fig. \ref{fig:tts}, we consider a two-time scale framework, where the large and small time scales are called frames and slots, respectively. The system's timeline is discretized into $J$ frames, denoted as $\mathcal{J}=\left\{1,2,\cdots,J\right\}$. Each frame $j \in \mathcal{J}$ is equally divided into $T$ time slots, denoted as $\mathcal{T}=\left\{1,2,\cdots,T\right\}$. In our considered system, at the beginning of each frame, the BS makes service caching decisions. At the beginning of each slot, the BS prices each program based on user demands, and followed by users determining the offloading proportions. It is assumed that all tasks can be completed before the end of the corresponding slot.

\subsection{Task Model}
Suppose that each user $m \in \mathcal{M}$ generates one computation task in each slot. The BS needs a service program to execute each task. The set of program types is $\mathcal{N}=\left\{1,\cdots,N\right\}$, where $N$ is the total number of types of service programs. We denote the task of user $m$ in slot $t$ as $\mathcal{I}_m^{t,j}=\left\{ \psi_m^{t,j},d_m^{t,j},\beta_m^{t,j},r_m^{t,j} \right\}$, where $\psi_m^{t,j} \in \mathcal{N}$ represents the service program type; $d_m^{t,j}$ represents the input data size; $\beta_m^{t,j}$ represents the computation intensity which is the number of CPU cycles per bit needed to execute the task; and $r_m^{t,j}=d_m^{t,j} \beta_m^{t,j}$ represents the total CPU cycles of the task. We assume that tasks can be divided into subtasks, users can offload $\alpha_m^{t,j} \mathcal{I}_m^{t,j}$ to the server, and $(1-\alpha_m^{t,j})  \mathcal{I}_m^{t,j}$ can be executed locally in parallel, where $\alpha_m^{t,j} \in [0,1]$ represents the offloading proportion \cite{diamanti2022incentive}.

\subsection{Service Caching Model}
The BS and all users have common prior knowledge of the estimated popularity  $\mathcal{Y}^{j}$ of all service programs based on the historical information, where $\mathcal{Y}^{j}=\left\{y_1^{j},y_2^{j},\cdots,y_N^{j}\right\}$, and each $y_n^j \in \mathcal{Y}^{j}$ denotes the probability of the type-$n$ program, i.e., $y_n^j \in [0,1]$, and we have $\sum_{n =1}^N y_{n}^{j}=1$. Specifically, we calculate the estimated popularity $y_{n}^{j}$ based on the numbers of task requests in the previous two frames:
\begin{equation}
	y_n^j=\frac{\nu_n^{j-1}+\nu_n^{j-2}}{\sum_{i = 1}^N \, \nu_i^{j-1}+\nu_i^{j-2}},
\end{equation}
where $\nu_n^j$ represents the number of tasks for the type-$n$ program in frame $j$. Moreover, the true popularity, which is publicly unknown, is defined as $\overline{y}_n^j=\frac{\nu_n^{j}}{\sum_{i=1}^N \, \nu_i^{j}}$.

The BS decides which programs to cache according to estimated popularity and its own caching capacity at the beginning of frame $j$. The caching decision is denoted as a binary indicator $x_{n}^{j}$, where $x_{n}^{j}=1$ if the type-$n$ program is cached at the BS in frame $j$, and $x_{n}^{j}=0$ otherwise. The set of caching results is denoted as $\mathbf{X}^{j} = \left\{x^{j}_{n}:\forall n \in  \mathcal{N} \right\}$. Moreover, the BS's caching decision is subject to the following capacity constraint:
\begin{equation}
	\sum_{n=1}^{N}x_{n}^{j}z_{n} \leq Z,
\end{equation}	
where $z_{n}$ is the size of the type-$n$ program and $Z$ is the caching capacity of the BS. Notably, user $m$ can offload its task to the server only if the BS pre-caches the corresponding program $\psi_m^{t,j}$ in frame $j$:
\begin{equation}
	\alpha_m^{t,j}\left(1-x_{\psi_m^{t,j}}^{j}\right)=0.
\end{equation}

\subsection{Communication Model}
In addition to the cached service programs, the BS needs the input data to execute the users' tasks. Suppose that users upload their input data to the edge server through the FDM or OFDM. Each user participating in offloading is fairly allocated the same bandwidth $W/M^{t,j}$, where $W$ is the available channel bandwidth, and $M^{t,j}$ is the number of offloading users, i.e., $M^{t,j}=\sum_{m = 1}^{M} \lceil \alpha_m^{t,j} \rceil$, where $ \lceil \alpha \rceil$ represents rounding up. Let $p_m^{t,j}$ and $h_m^{t,j}$ denote the transmit power and the wireless channel gain between user $m$ and the BS, respectively. Specifically, $h_m^{t,j}=\lambda \xi_m^{t,j} (o_m^{t,j})^{-e}$, where 
$\lambda$ is the pathloss constant, $\xi_m^{t,j} \sim E(1)$ is the small-scale fading channel power gain, $o_m^{t,j}$ is the distance between the BS and user $m$, and $e$ is the pathloss exponent \cite{gu2020deep}. Moreover, we assume that there is noise with zero mean and identical variance $\sigma^2$ for all users. The uplink data rate from user $m$ to the BS is
\begin{equation}
	R_m^{t,j}=\frac{R^{t,j}}{M^{t,j}}=\frac{W}{M^{t,j}}\log_2\left(1+\frac{p_m^{t,j}h_m^{t,j}}{\sigma^2}\right).
\end{equation}
As a result, we can define the uplink transmission delay of user $m$ when offloading its input data to the BS:
\begin{equation}
	D_{m,t,j}^\text{tra}=\frac{d_{m,t,j}^{\text{off}}}{R_m^{t,j}}=\frac{\alpha_m^{t,j} d_m^{t,j}}{R_m^{t,j}}.
\end{equation}
After the BS finishes executing the offloaded tasks, users need to download the results from the BS and combine them with the locally calculated results to obtain the final task results. However, the amount of data representing the computation results is usually much smaller than the amount of input data for the tasks, so we ignore the download transmission delay in our model.

\subsection{Computation Model}
\subsubsection{Local Computation}
The computation delay spent on the local execution of a task is defined as:
\begin{equation}
	D_{m,t,j}^\text{loc}=\frac{r_{m,t,j}^\text{loc}}{f_m}=\frac{(1-\alpha_m^{t,j}) r_m^{t,j}}{f_m},
\end{equation}
where $f_m$ is the CPU computation frequency of user $m$.
\subsubsection{Offloading Computation}
When the BS receives multiple offloaded tasks, the computation frequency $F$ of the server is equally divided among the users. This means that the greater the number of users that participate in MEC offloading, the fewer computation resources they will be allocated and the longer execution delay will be experienced:
\begin{equation}
	D_{m,t,j}^\text{exe}=\frac{r_{m,t,j}^{\text{off}}}{F^{t,j}}=\frac{\alpha_m^{t,j} r_m^{t,j}}{F/M^{t,j}}.
\end{equation}

\section{Problem Formulation}
\subsection{Utility Function of the BS}
\subsubsection{Large Time Scale}
Based on the dynamic popularity of programs, the BS needs to selectively cache appropriate programs to increase its profits. Suppose that the payment in slot $t$ is defined as $U_{t,j}^{\text{BS}}$. The utility function of the BS for the whole time frame $j$ is defined as the difference between the payments for all time slots within the time frame and the caching costs paid to the program provider:
\begin{equation}
	\begin{aligned}
		U_j^{\text{BS}}\left(\mathbf{X}^{j}\right) &= \sum_{t = 1}^T U_{t,j}^{\text{BS}} - \sum_{n =1}^N (1-x_{n}^{j-1})x_{n}^{j} \omega_{n}^{j},\\
	\end{aligned} \label{utility function of frame T}
\end{equation}
where $\omega_{n}^{j}$ represents the caching cost for the type-$n$ program. In particular, for each frame $j$, the BS does not need to obtain new service data for the type-$n$ program if it is already cached in the previous frame.

At the beginning of each time frame, the BS must determine which services to cache based on the estimated popularity $\mathcal{Y}^j$, previous caching results, and potential future task offloading requests within the bounds of the constraints. The optimization problem for the BS at the large time scale is formulated as follows:
\begin{equation}
	\begin{aligned}
		(P1) \;	\max_{\mathbf{X}^{j}}& &&U_j^{\text{BS}}\left(\mathbf{X}^{j}\right), \\
		\text{s.t.} & && \displaystyle\sum_{n=1}^{N} x_{n}^{j}z_{n} \leq Z, &&\\
		& &&x_{n}^{j} \in \left\{0,1\right\}, &&\forall n \in \mathcal{N},\\
		& &&\omega_{n}^{j}x_{n}^{j-1}=0, &&\forall n \in \mathcal{N}.
	\end{aligned}
\end{equation}

\subsubsection{Small Time Scale}
Given the caching result $\mathbf{X}^{j}$ in frame $j$, the BS aims to maximize its profit by adjusting the pricing strategy for each slot $t$. Suppose that the BS charges $\pi_{n}^{t,j}$ per CPU cycle from users when executing type-$n$ offloading tasks. The payments received from users in slot $t$ are defined as:
\begin{equation}
	U_{t,j}^{\text{BS}}\left(\bm{\Pi}^{t,j},\bm{\alpha}^{t,j}\right) = \sum_{m = 1}^M \alpha_m^{t,j} r_m^{t,j} \pi_{\psi_m^{t,j}}^{t,j} x_{\psi_m^{t,j}}^{j}, \label{utility function of BS}
\end{equation}
where $\bm{\Pi}^{t,j}=\left\{\pi^{t,j}_{n}:\forall n \in  \mathcal{N}\right\}$ is the price set of all programs and $\bm{\alpha}^{t,j}=\left\{\alpha^{t,j}_{m}:\forall m \in \mathcal{M}\right\}$ is the offloading decision set of all users. 

At the beginning of each time slot, the BS must make pricing decisions considering users' offloading behaviors.
The optimization problem for the BS at the small time scale is formulated as follows:
\begin{equation}
	\begin{aligned}
		(P2) \;\max_{\bm{\Pi}^{t,j}}& && U_{t,j}^{\text{BS}}\left(\bm{\Pi}^{t,j},\bm{\alpha}^{t,j}\right), \\
		\text{s.t.}& && \pi^{t,j}_{n} \geq 0, && \forall n \in \mathcal{N}.
	\end{aligned}
\end{equation}

\subsection{Cost Function of Users}
Each user strategically determines an optimal offloading proportion to minimize its total cost, characterized by the weighted sum of the payment to the BS and the time delay. In particular, offloading and local computation for the user are performed in parallel during the completion lifetime of the task, and the task is not complete until all the results are received. Therefore, the time delay is defined as the maximum value between the offloading time delay $D_{m,t,j}^\text{off}$ and the local computation time delay $D_{m,t,j}^\text{loc}$, where $D_{m,t,j}^\text{off}$ consists of the transmission delay $D_{m,t,j}^{\text{tra}}$ and the offloading computation delay $D_{m,t,j}^\text{exe}$:
\begin{equation}
	\begin{aligned}
		D_m^{t,j} &= \max\left\{D_{m,t,j}^{\text{tra}}+D_{m,t,j}^\text{exe},D_{m,t,j}^{\text{loc}}\right\}. \label{time delay}
	\end{aligned}
\end{equation}
Based on this, the cost function of user $m$ is defined as the weighted sum of the payment and time delay:
\begin{equation}
	\begin{aligned}
		C_{m,t,j}^{\text{user}}\left(\bm{\Pi}^{t,j},\alpha_m^{t,j},\bm{\alpha}_{-m}^{t,j}\right) &= \alpha_m^{t,j}r_m^{t,j} \pi_{\psi_m^{t,j}}^{t,j} x_{\psi_m^{t,j}}^{j} + \theta  D_m^{t,j},  \label{cost function of user}
	\end{aligned}
\end{equation}
where $ \bm{\alpha}_{-m}^{t,j} = \left\{\alpha_1^{t,j},\cdots,\alpha_{m-1}^{t,j},\alpha_{m+1}^{t,j},\cdots,\alpha_M^{t,j}\right\}$ and $\theta$ is the price adjustment coefficient. 

After the BS announces the prices, user $m$ must determine its offloading proportion considering the prices and other users' offloading behaviors. The optimization problem for user $m$ is formulated as:
\begin{equation}
	\begin{aligned}
		(P3) \; \min_{\alpha_m^{t,j}} & && C_{m,t,j}^{\text{user}}\left(\bm{\Pi}^{t,j},\alpha_m^{t,j},\bm{\alpha}_{-m}^{t,j}\right), \\
		\text{s.t.} & && \alpha_m^{t,j} \in [0,1],  && \forall m \in \mathcal{M},\\
		& &&\alpha_m^{t,j}\left(1-x_{\psi_m^{t,j}}^{j}\right)=0, &&\forall m \in \mathcal{M}.
	\end{aligned}
\end{equation}

\section{Popularity-Aware Dynamic Service Caching for the Large Time Scale}
In this section, we primarily investigate service caching for the large time scale. We start by analyzing the complexity of  problem (P1) and identifying the challenges in sloving it. Then, we propose the GNDRL algorithm based on the MDP and provide specific details on how it addresses the issues related to service caching.	

\begin{figure}[tbp]
	\vspace{-0.4cm}
	\centering
	\includegraphics[width=0.42\textwidth]{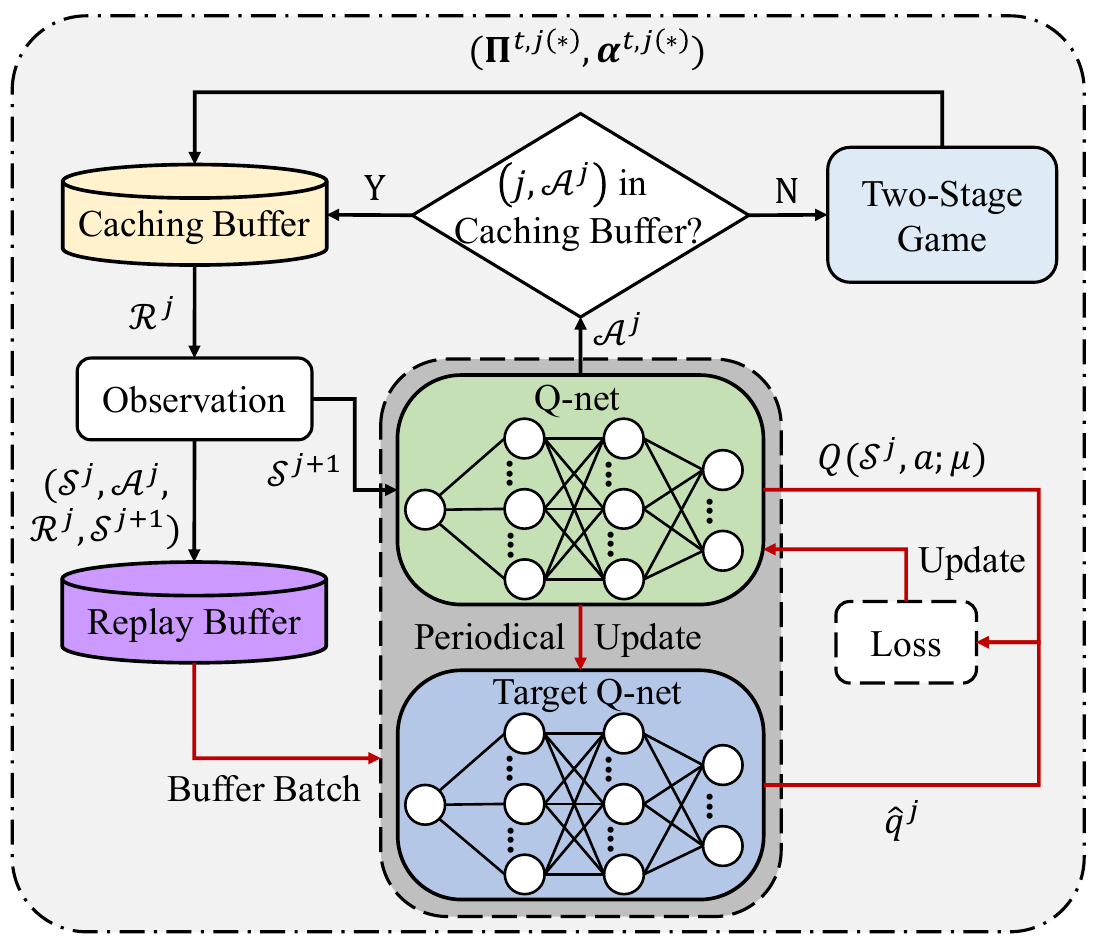}
	\caption{The framework of GNDRL.}
	\label{fig:GNDRL}
	\vspace{-0.2cm}
\end{figure}

Solving (P1) primarily presents the following challenges. First, (P1) is temporally coupled across the large time scale due to the caching cost $\omega_{n}^{j}$. Second, (P1) is a 0-1 integer nonlinear programming problem that presents a considerable computation challenge even if future information is known in advance. However, the BS cannot accurately predict the tasks' details in subsequent time slots, including the types of programs, size of the input data, or number of CPU computation cycles. A closely related problem has been demonstrated to be NP-hard, as shown in \cite{ren2022adaptive} and \cite{poularakis2019joint}. To overcome the above challenges, we propose the GNDRL algorithm to solve the service caching problem as shown in Fig. \ref{fig:GNDRL}.

The MDP for GNDRL, denoted as $\langle \mathcal{S},\mathcal{A},\mathcal{R} \rangle $, is expressed as follows.
\subsubsection{State} 
At the beginning of each frame, the observations made by the BS regarding the environment consist only of the estimated popularity information and the caching results from the previous frame. Therefore, the state in frame $j$ is defined as:
\begin{equation}
	\mathcal{S}^{j}=\left\{j,y_1^{j},\cdots,y_N^{j},\mathcal{A}^{j-1}\right\}.
\end{equation}
\subsubsection{Action} 
The BS makes service caching decisions based on the state information. Given that $x_{n}^{j} \in \left\{0,1\right\}$, we can indeed consider the set $\mathbf{X}^{j}$ as a binary number. To reduce the dimension of the state space and action space, we define the action in frame $j$ as the decimal form of the binary number:
\begin{equation}
	\mathcal{A}^{j}= \sum_{n=1}^N 2^{(n-1)x_{n}^{j}},
\end{equation}
where the value range of the decimal number is $[0,2^N-1]$.
\subsubsection{Reward}
After the caching results are announced, for each slot $t$, there formulates a two-stage game between the BS and users, which will be described in Section \ref{sec:stackelberg}. By analyzing the game, we can obtain the payments and caching costs of the BS. We define the reward as the utility function in frame $j$:
\begin{equation}
	\mathcal{R}^{j} = U_{j}^{\text{BS}},
\end{equation}
if the caching result $\mathbf{X}^{j}$ satisfies the caching capacity constraints; otherwise, inspired by the Karush-Kuhn-Tucker (KKT) conditions, we introduce a penalty factor $\rho_c$: $\mathcal{R}^{j} = - \rho_c (\sum_{n=1}^{N} x_{n}^{j}z_{n} - Z)$.

\normalem
\begin{algorithm}[t]
	\caption{GNDRL-Based Service Caching}\label{GNDRL}
	\KwIn{State space $\{\mathcal{S}^1,\cdots,\mathcal{S}^{T}\}$}
	\KwOut{Reward $\{\mathcal{R}^1,\cdots,\mathcal{R}^{T}\}$, Service caching $\{\mathbf{X}^1,\cdots,\mathbf{X}^{T}\}$}
	Initialize replay buffer $B_r$\;
	Initialize caching buffer $B_c$ with initial value -2\;
	Initialize Q-net $Q$ with random weights $\mu$\;
	Initialize target Q-net $\hat{Q}$ with weights $\hat{\mu} = \mu$\;
	
	\For{episode = $1,2,\cdots,E$}{
		Initialize action $\mathcal{A}^0 \gets 0$\;
		\For{$T = 1,2,\cdots,T$}{
			Choose action $\mathcal{A}^j$ with (\ref{epsilon greedy})\;
			\eIf{$(T,\mathcal{A}^j) \in  B_c$}{
				$\mathcal{R}^j \gets B_c(T,\mathcal{A}^j)$\;
			}{
				Translate action $\mathcal{A}^j$ into caching result $\mathbf{X}^j$\;
				Calculate $\mathcal{R}^j$ using (\ref{utility function of frame T}) \;
				$B_c(T,\mathcal{A}^j) \gets \mathcal{R}^j$\;
			}
			Observe next state $\mathcal{S}^{j+1}$\;
			Store transition ($\mathcal{S}^j,\mathcal{A}^j,\mathcal{R}^j,\mathcal{S}^{j+1}$) into $B_r$\; 
			Sample random minibatch of transitions ($\mathcal{S}^j,\mathcal{A}^j,\mathcal{R}^j,\mathcal{S}^{j+1}$) from $B_r$\;
			Obtain target q-value $\hat{q}^j$ with (\ref{target q value})\;
			Update $\mu$ through backpropagation according to (\ref{MSE}) with learning rate $\rho_l$\;
			Every $\kappa$ steps  update $\hat{\mu} \gets \mu$.
		}
	}
\end{algorithm}
\ULforem

	The algorithmic process of GNDRL is shown in Algorithm \ref{GNDRL}. After initialization, given state $\mathcal{S}^{j}$ for each frame $j$, the BS obtains an action $\mathcal{A}^{j}$ based on the $\epsilon$-greedy strategy:
\begin{equation}
	\mathcal{A}^{j} = \begin{cases}
		\text{Random}\left(0,2^{N}-1\right),&\text{with probability $\epsilon$},\\
		\argmax_{a}Q\left(\mathcal{S}^{j},a;\mu\right),&\text{otherwise}.
	\end{cases} \label{epsilon greedy}
\end{equation}
Then, based on game analysis and the caching result of the previous time frame, the BS obtains the corresponding reward $\mathcal{R}^{j}$ while concurrently observing the next state $\mathcal{S}^{j+1}$. Subsequently, the BS stores transition ($\mathcal{S}^{j},\mathcal{A}^{j},\mathcal{R}^{j},\mathcal{S}^{j+1}$) into replay buffer $B_r$. The parameter $\mu$ update process for the Q-net $Q$ is as follows: the BS randomly samples a minibatch of transitions, and based on the sampling results, the target q-value can be obtained as:
\begin{equation}
	\hat{q}^{j} = \begin{cases}
		\mathcal{R}^{j},&\text{for terminal $J$},\\
		\mathcal{R}^{j}+ \gamma \max_{a'}\hat{Q}\left(\mathcal{S}^{j+1},a';\hat{\mu}\right),&\text{otherwise}.
	\end{cases} \label{target q value}
\end{equation}
Then, based on the mean squared error (MSE) between the q-value and target q-value, the parameter $\mu$ is updated through back propagation with a learning rate $\rho_l$:
\begin{equation}
	\text{MSE}\left(q^{j},\hat{q}^{j}\right) = \left(\hat{q}^{j}-Q\left(\mathcal{S}^{j},\mathcal{A}^{j};\mu\right)\right)^2. \label{MSE}
\end{equation}
Moreover, the BS periodically updates the parameters $\hat{\mu}$ of the target Q-net.

In particular, we note that for a certain frame $j$, the same action $\mathcal{A}^{j}$ leads to the same reward. To avoid redundant computation, we design a caching buffer $B_c$ as a hash-map where each combination of time frames and actions ($j,\mathcal{A}^{j}$) is mapped to a corresponding reward. Whenever the BS observes a state $\mathcal{S}^{j}$ and selects an action $\mathcal{A}^{j}$, it checks whether the tuple ($j,\mathcal{A}^{j}$) was computed in a previous episode by consulting the buffer. If the result is available, the BS can directly observe $\mathcal{R}^{j}$ in the buffer; otherwise, it iterates over all slots $t$ to compute the reward and stores it in $B_c$ for future reference.

\section{Pricing and Task Offloading For the small Time Scale} 
\label{sec:stackelberg}
In this section, we analyze pricing and task offloading for the small time scale. We model the competitive interaction between the BS and users as a two-stage Stackelberg game and backward induction is utilized to analyze the game. By modeling a selfish offloading subgame, we derive an optimal threshold-based offloading strategy under incomplete information. Subsequently, we propose two pricing strategies  based on the users' best responses. Finally, we validate the existence of the Nash equilibrium (NE) of the subgame and the Stackelberg equilibrium (SE) of the Stackelberg game.
\subsection{Two-Stage Stackelberg Game Formulation}
\begin{figure}[t]
		\vspace{-0.4cm}
		\centering
		\includegraphics[width=0.42\textwidth]{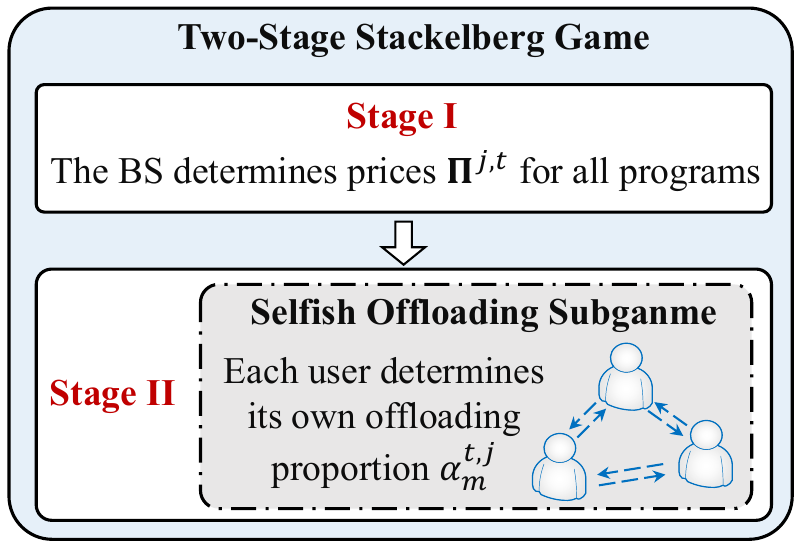}
		\caption{Two-stage Stackelberg game.}
		\label{fig:Stackelberg}
		\vspace{-0.2cm}
	\end{figure}

For the small time scale, the BS first sets prices for each program at the beginning of slot $t$. Afterwards, users need to pay the BS for computation services when offloading their tasks. Considering the payment and the time delay, each user determines its own offloading proportion $\alpha_m^{t,j}$ according to the prices. The behaviors of the BS and users interact with each other, which makes this problem hard to optimize. Thus, we model the competitive interaction as a one-leader, multi-follower Stackelberg game as shown in Fig. \ref{fig:Stackelberg}.

\vspace{-0.1cm}
\begin{game}[Stackelberg Game between the BS and Users]
\end{game}    
\vspace{-0.2cm}
\begin{itemize}
	\item \textit{\textbf{Players:}} The BS acts as the leader, and users act as the followers.
	\item \textit{\textbf{Strategies:}} The BS determines the prices of each program $n \in \mathcal{N}$ in Stage \Rmnum{1}, and each user $m \in \mathcal{M}$ observes the program prices and determines the offloading proportion in Stage \Rmnum{2}. 
	\item \textit{\textbf{Objectives:}} The BS aims to maximize its payments $ U_{t,j}^{\text{BS}}$, and each user aims to minimize its total cost $ C_{t,j}^{\text{user}}$.
\end{itemize}
\vspace{0.1cm}

	In what follows, we analyze the Stackelberg game using backward induction: we first establish a resource competition subgame between users in Stage \Rmnum{2}, and derive the optimal offloading proportions for all users. Then, we optimize the pricing of the BS in Stage \Rmnum{1}, by predicting the users' best response behaviors.

\subsection{Stage \Rmnum{2}: Users' Offloading Proportion }
According to the definition of $M^{t,j}$, the offloading strategies of other self-interested users, i.e., $\bm{\alpha}_{-m}^{t,j}$, affect the available bandwidth $\frac{R^{t,j}}{M^{t,j}}$ and computation resource $\frac{F}{M^{t,j}}$ for each user, thereby impacting the total cost of user $m$. (P3) is formulated under the assumption that the user has knowledge of $\bm{\alpha}_{-m}^{t,j}$. However, in real-world scenarios, users are generally reluctant to share their personal information with other users, e.g., their computation frequency $f_m$, task detail $\mathcal{I}_m^{t,j}$ and offloading proportion $\alpha_m^{t,j}$. Consequently, the specific offloading number $M^{t,j}$ is unknown to all users in slot $t$. We model the mutual competition between all users under incomplete information as a selfish offloading subgame.
\vspace{-0.1cm}
	\begin{game}[Selfish Offloading Subgame between Users]
	\end{game}    
	\vspace{-0.2cm}
	\begin{itemize}
		\item \textit{\textbf{Players:}} The set $\mathcal{M}$ of users.
		\item \textit{\textbf{Strategies:}} By observing the prices, each user $m \in \mathcal{M}$ determines its offloading proportion considering mutual competition from other users.
		\item \textit{\textbf{Objectives:}} Each user aims to minimize its total cost $ C_{t,j}^{\text{user}}$.
		\end{itemize}
	\vspace{0.1cm}

To derive the NE of Game 2, we first try to derive the offloading strategy under complete information, where each user knows the number of offloading users $M^{t,j}$.

By observing the prices announced by the BS in Stage \Rmnum{1}, we can deduce  (\ref{time delay}) to obtain
\begin{equation}
	D_m^{t,j}=\begin{cases}
		D_{m,t,j}^{\text{tra}}+D_{m,t,j}^{\text{exe}},& \delta_m^{t,j} \leq \alpha_m^{t,j} \leq 1, \\
		D_{m,t,j}^{\text{loc}},& 0 \leq \alpha_m^{t,j} < \delta_m^{t,j}, \label{time delay 0}
	\end{cases}
\end{equation}
where
\begin{equation}
	\delta_m^{t,j} = \frac{\frac{r_m^{t,j}}{f_m}}{\frac{d_m^{t,j}}{R_m^{t,j}}
		+\frac{r_m^{t,j}}{F^{t,j}}+\frac{r_m^{t,j}}{f_m}}.
\end{equation}
Substituting (\ref{time delay 0}) into (\ref{cost function of user}), we can obtain:
\begin{equation}
	C_{m,t,j}^{\text{user}}=\begin{cases}
		\alpha_m^{t,j}r_m^{t,j} \pi_{\psi_m^{t,j}}^{t,j} x_{\psi_m^{t,j}}^{j} + \theta  D_{m,t,j}^{\text{off}},& \delta_m^{t,j} \leq \alpha_m^{t,j} \leq 1, \\
		\alpha_m^{t,j}r_m^{t,j} \pi_{\psi_m^{t,j}}^{t,j} x_{\psi_m^{t,j}}^{j} + \theta  D_{m,t,j}^{\text{loc}},& 0 \leq \alpha_m^{t,j} < \delta_m^{t,j}. \label{cost function of user 1}
	\end{cases}
\end{equation}
Calculating the minimum value of (\ref{cost function of user 1}) yields the optimal offloading proportion of user $m$, which is a step function of the price $\pi_{\psi_m^{t,j}}^{t,j}$:
\begin{equation}
	\alpha_m^{t,j(*)}\left(\pi_{\psi_m^{t,j}}^{t,j}\right)=\begin{cases}
		0,&\pi_{\psi_m^{t,j}}^{t,j}-\frac{\theta}{f_m} < 0,\\
		\delta_m^{t,j},&\pi_{\psi_m^{t,j}}^{t,j}-\frac{\theta}{f_m} \geq 0,
	\end{cases} \label{optimal proportion}
\end{equation}
where $\frac{\theta}{f_m}$ is defined as the \textit{characteristic parameter} of user $m$.

Now, we turn to the incomplete information scenario where the offloading number $M^{t,j}$ is not publicly known. In particular, each user is unaware of the exact values of the allocated uplink rate $\frac{R^{t,j}}{M^{t,j}}$ and  computation frequency $\frac{F}{M^{t,j}}$.
	
\begin{assumption}[Uniform Distribution]
	The frequency of users is independent and identically distributed with a probability density function (PDF) $g(\cdot)$ and a cumulative distribution function (CDF) $G(\cdot)$. The distribution function is common prior knowledge for the BS and all users \cite{yan2021pricing}.
\end{assumption}
\begin{lemma}
	Each user can estimate the same number of offloading participants $\hat{M}^{t,j}$ under incomplete information:
	\begin{equation}
		\hat{M}^{t,j} = 1+ \left(M-1\right)\times \sum_{n =1}^N x_{n}^{j} y_{n}^{j} G\left(\frac{\theta}{\pi_{n}^{t,j}}\right). \label{estimated number}
	\end{equation} 
	\label{lemma:offnum}
\end{lemma}
\vspace{-0.1cm}
\begin{proof}
	On the basis of (\ref{optimal proportion}), user $m$ will offload its task to the edge server if $f_{m} \leq \theta/\pi_{\psi_m^{t,j}}^{t,j}$. Therefore, under the scenario of incomplete information, each user estimates the probabilities of other users participating in offloading  as following a binomial distribution 
	$B(M-1,G(\frac{\theta}{\pi_n^{t,j}}))$, and the expectation of this distribution is $(M-1)G(\frac{\theta}{\pi_n^{t,j}})$. In addition, without knowing the specific program types of other users, user $m$ can only estimate the probability of other users requesting type-$n$ tasks based on the program popularity $y_n^j$. From the perspective of user $m$, the number of other users expected to participate in offloading is $(M-1)\times \sum_{n=1}^{\mathcal{N}} x_n^j y_n^j G(\frac{\theta}{\pi_n^{t,j}})$.
\end{proof}

Substituting  (\ref{estimated number}) into $\delta_m^{t,j}$ in (\ref{optimal proportion}), we obtain the optimal threshold-based offloading strategy $\alpha_m^{t,j(*)}$ for user $m$ under incomplete information in Stage \Rmnum{2}.

\subsection{Stage \Rmnum{1}: BS's Pricing Strategy}
\label{section:pricing}
Although users do not know each other's private information, they will share these details with the BS for task offloading requirements. Therefore, knowing the users' best responses to different prices in Stage \Rmnum{2}, the BS can optimize its utility function by predicting the users' offloading behaviors.
We define a set $\mathcal{M}_{n}^{t,j}=\left\{m:\psi_m^{t,j} = j,\forall m \in \mathcal{M}\right\}$ that contains the users who need to compute type-$n$ tasks, and $|\mathcal{M}_{n}^{t,j}|$ is the size of this set. By substituting (\ref{optimal proportion}) into (\ref{utility function of BS}), we can divide (\ref{utility function of BS}) into different service programs for discussion, i.e., $
	U_{t,j}^{\text{BS}} = \sum_{n =1}^N U_{n,t,j}^{\text{BS}}, \label{utility function of BS 1}
	$
where 
\begin{equation}
	U_{n,t,j}^{\text{BS}}\left(\pi_{n}^{t,j}\right) = \sum_{m \in \mathcal{M}_{n}^{t,j}} \alpha_m^{t,j(*)}\left(\pi_{n}^{t,j}\right) r_m^{t,j} \pi_{n}^{t,j} x_{n}^{j} \label{utility function of program}
\end{equation}
is the profit of providing computing service of the type-$n$ program in slot $t$.

The step function $\alpha_m^{t,j(*)}(\pi_n^{t,j})$ in (\ref{optimal proportion}) makes the function (\ref{utility function of program}) discontinuous and non-differentiable and it is difficult to obtain a closed-form solution. Therefore, by analyzing the function characteristics (algorithm profitability) or using an approximate function (computation complexity), we propose two pricing algorithms respectively: characteristic parameter traversal pricing optimization (CPTO) and solver-based continuous approximation pricing optimization (SCAO). Detailed descriptions of these algorithms follow.

\SetArgSty{textup}	
\normalem	
\begin{algorithm}[t]
	\caption{CPTO in Stage \Rmnum{1}}\label{CPTO}
	\KwIn{$\hat{M}^{t,j}$}
	\KwOut{$\bm{\Pi}^{t,j(*)} = \{\pi_n^{t,j(*)}: \forall j \in \mathcal{N} \}$}
	\For{$j \in \mathcal{N}$}{
		$\pi_n^{t,j(*)} \gets 0 $\;
		\If{$x_n^j$ = 1}{
				\For{$m \in \mathcal{M}_n^{t,j}$}{
				\If{$U_{n,t,j}^{\text{BS}}\left(\frac{\theta}{f_{m}}\right) >$ \textit{MaxValue}}{
					\textit{MaxValue} $\gets$ $U_{n,t,j}^{\text{BS}}\left(\frac{\theta}{f_{m}}\right)$\;
					$\pi_n^{t,j(*)} \gets \frac{\theta}{f_{m}}$\;
				}
			}
		}
	}		
\end{algorithm}

\normalem	
\begin{algorithm}[t]
	\caption{SCAO in Stage I}\label{SCAO}
	\KwIn{$\hat{M}^{t,j},\mathbb{S}$}
	\KwOut{$\bm{\Pi}^{t,j(*)} = \{\pi_n^{t,j(*)}: \forall j \in \mathcal{N} \}$}
	\For{$j \in \mathcal{N}$}{
		$\pi_n^{t,j(*)} \gets 0 $\;
		\If{$x_n^j$ = 1}{
		Initialize temporary parameter \textit{MaxValue} $\gets$ 0\;
		\For{$\textit{StartPoint} \in \mathbb{S}$}{
			\textit{Zero} $\gets$ \text{SolveZero}($\nabla \tilde{U}_{n,t,j}^{\text{BS}}$, \textit{StartPoint})\;
			\If{$U_{n,t,j}^{\text{BS}}(\textit{Zero}) >$ \textit{MaxValue}}{
				\textit{MaxValue} $\gets$ $U_{n,t,j}^{\text{BS}}(\textit{Zero})$\;
				$\pi_n^{t,j(*)}  \gets$ \textit{Zero}\;
			}
		}
	}

	}
\end{algorithm}
\ULforem

\subsubsection{CPTO} We observe from (\ref{optimal proportion}) that the offloading strategy of user $m$ is related to its characteristic parameters $\frac{\theta}{f_{m}}$, based on which we propose CPTO to find the optimal pricing strategy $\bm{\Pi}^{t,j(*)}$ in Stage \Rmnum{1}.
\begin{lemma}
	There exists a maximum value of (\ref{utility function of program}), and the corresponding optimal price $\pi_n^{t,j(*)}$ is equal to the characteristic parameter $\frac{\theta}{f_{m}}$ of a certain user $m \in \mathcal{M}_n^{t,j}$. \label{lemma:CPTO}
\end{lemma}
\vspace{0.1cm}
\begin{proof}
	When $\pi_n^{t,j} = 0$, obviously we have $U_{t,j}^{\text{BS}} = 0$. Without loss of generality, we assume that $\mathcal{M}_n^{t,j}$ are arranged in ascending order according to the characteristic parameters of the users, i.e., $\frac{\theta}{f_{m-1}} \leq \frac{\theta}{f_{m}}, \forall m \in \mathcal{M}_n^{t,j}$. On the basis of (\ref{optimal proportion}), when $\pi_n^{t,j} >  \frac{\theta}{f_{|\mathcal{M}_n^{t,j}|}}$, no user will offload its task to the BS; thus, we also have $U_{t,j}^{\text{BS}} = 0$. In addition, $U_{t,j}^{\text{BS}}$ is monotonically increasing in the interval $(\frac{\theta}{f_{m-1}},\frac{\theta}{f_{m}}]$, which means that given a price $\pi \in (\frac{\theta}{f_{m-1}},\frac{\theta}{f_{m}}]$, we have $U_{t,j}^{\text{BS}}(\pi) \leq U_{t,j}^{\text{BS}}(\frac{\theta}{f_{m}}) $, and $U_{t,j}^{\text{BS}}(\frac{\theta}{f_{m}})$ is a finite value.
Therefore, there must exist an upper bound of $U_{t,j}^{\text{BS}}$, and the upper bound corresponds to the characteristic parameter of a certain user .
\end{proof}

On the basis of Lemma \ref{lemma:CPTO}, we design the algorithmic procedure of CPTO in Algorithm \ref{CPTO}.

\normalem
\begin{algorithm}[t]
	\caption{Stackelberg Equilibrium in Slot $t$}\label{SE of slot}
	\KwIn{$\mathbf{X}^j$}
	\KwOut{($\bm{\Pi}^{t,j(*)},\bm{\alpha}^{t,j(*)}$)}
	Initialize the prices $\bm{\Pi}^{t,j(0)} = \{\pi_n^{t,j(0)}: \forall j \in \mathcal{N} \}$\;
	$k \gets 1$\;
	\Repeat{
		$\bm{\Pi}^{t,j(*)}$ \text{converges}
	}{
		
		Estimate $\hat{M}^{t,j(k-1)}$ given prices $\bm{\Pi}^{t,j(k-1)}$\;
		\For{$j \in \mathcal{N}$}{
			Calculate the price $\pi_n^{t,j(k)}$ through CPTO or SCAO given the estimated number $\hat{M}^{t,j(k-1)}$\;
		}
		$k \gets k+1$\;
	}
\end{algorithm}
\ULforem

\subsubsection{SCAO}
Although leveraging CPTO to optimize pricing has the potential to yield substantial profits for the BS, it incurs long computation delays when $M$ or $N$ is large. Therefore, we propose another low-complexity algorithm, SCAO, which prioritizes high computation efficiency at the cost of a certain amount of profit.

As mentioned above, the difficulty of solving this problem lies in the step function $\alpha_m^{t,j(*)}(\pi_n^{t,j})$. We use a sigmoid function 
\begin{equation}
	\tilde{\alpha}_m^{t,j(*)}\left(\pi_{\psi_m^{t,j}}^{t,j}\right) = \delta_m^{t,j} \times \text{sigmoid}\left(\frac{\theta}{f_{m}}-\pi_{\psi_m^{t,j}}^{t,j}\right) 
\end{equation}
to approximate (\ref{optimal proportion}), which makes the utility function continuous and differentiable:
\begin{equation}
	\tilde{U}_{n,t,j}^{\text{BS}}(\pi_n^{t,j}) = \sum_{i \in \mathcal{M}_n^{t,j}}  r_m^{t,j}   \delta_m^{t,j}  \pi_n^{t,j} \times \text{sigmoid}\left(\frac{\theta}{f_{m}}-\pi_n^{t,j}\right). \label{utility function of SCAO}
\end{equation}
Taking the derivative of (\ref{utility function of SCAO}), we can obtain
\begin{equation}
	\begin{alignedat}{2}
		\nabla \tilde{U}_{n,t,j}^{\text{BS}} (\pi_n^{t,j}) &= &&\frac{d \tilde{U}_{n,t,j}^{\text{BS}} }{d \pi_n^{t,j}} \\
		&= &&\sum_{m \in \mathcal{M}_n^{t,j}}  r_m^{t,j}   \delta_m^{t,j} \times \text{sigmoid}\left(\frac{\theta}{f_{m}}-\pi_n^{t,j}\right) \\
		&&&\times \left(1-\pi_n^{t,j}\left(1-\text{sigmoid}\left(\frac{\theta}{f_{m}}-\pi_n^{t,j}\right)\right)\right).
	\end{alignedat}
	\label{first derivative of SCAO}
\end{equation}
Given the analytical expression of the first derivative, we can find the extremum points by solving for all its zeros. However, since (\ref{first derivative of SCAO}) contains non-linear terms such as the sigmoid function and summation, obtaining a closed-form analytical solution is almost impossible. Therefore, we consider a set of starting points $\mathbb{S}$ and use an iterative solver \text{SolveZero} to search for different zeros. By comparing the function values, we can determine the final pricing result.

\begin{table}[t]
	\renewcommand{\arraystretch}{1.05}
	\centering
	\captionsetup{labelsep=newline} 
	\caption{Parameter Setup}
	\label{Tab 1} 
	\begin{tabular}{c@{\hspace{.8ex}}l@{\hspace{.8ex}}c}
		\hline
		\multicolumn{1}{c}{Parameter} & \multicolumn{1}{c}{Description} & \multicolumn{1}{c}{Setting} \\

		\hline
		$W$        & Bandwidth        & 2 MHz        \\
		$p_m^{t,j}$       & Transmit power        & [80,200] mW       \\
		$\lambda$       & Pathloss constant        & 1   \\
		$e$       & Pathloss exponent        & 2       \\
		$\xi_m^{t,j} $       & Small-scale fading channel power gain        & E(1)      \\
		$o_m^{t,j}$       & Distance        & [100,1000] m       \\
		$\sigma^2$       & Gaussian noise power density       & $10^{-10} W$      \\
		
		$d_m^{t,j}$       & Input data size     & [200,1000] KB       \\
		$\beta_m^{t,j}$       & Computaton intensity       & [800,2000] cycles/bit        \\
		$f_m$       & Computation frequency of user $m$       & [0.5,4] MHz       \\
		$\theta$       & Price adjustment coefficient       & $2\times10^7$        \\
		$N$       & Number of programs        & 4        \\
		$\gamma$  & Discount factor        & 0.9  \\
		$\rho_c$  & Penalty factor       & 0.02  \\
		$\rho_l$  & Learning rate       & 0.001  \\
			
			$J$ 			      & Number of frames        & 5        \\
			$T$        & Number of slots        & 5 \\
					& Batch size      & 64 \\
		\hline
	\end{tabular}
		\vspace{-0.4cm}
\end{table}

\subsection{Analysis of Equilibrium}

We can define the NE of Game 2 as follows: 
if the subgame reaches an NE, no user can alter its offloading proportion to reduce the total costs.
\begin{definition}[Nash Equilibrium of Game 2]
	A proportion profile $\bm{\alpha}^{t,j(*)}$ is an NE in Stage \Rmnum{2} if for each user $m \in \mathcal{M} $:
	\begin{equation}
		C_{m,t,j}^{\text{user}}(\alpha_{m}^{t,j(*)},\bm{\alpha}_{-m}^{t,j(*)}) \leq C_{m,t,j}^{\text{user}}(\alpha_m^{t,j},\bm{\alpha}_{-m}^{t,j(*)}).
	\end{equation}
\end{definition}

The NE $\bm{\alpha}^{t,j(*)} (\bm{\Pi}^{t,j})$ is the best response of the users to the pricing scheme. Differences in pricing may lead to changes in the NE. A stable system under a Stackelberg game often reaches an SE, which means that neither the leader BS nor a follower user can obtain greater utility or bear lower costs by adjusting its own strategy.

\begin{definition}[Stackelberg Equilibrium of Game 1]
	A combination $\left(\bm{\Pi}^{t,j(*)},\bm{\alpha}^{t,j(*)}\right)$ is an SE if
	\begin{equation}
		\begin{aligned}
			U_{t,j}^{\text{BS}}(\bm{\Pi}^{t,j(*)},\bm{\alpha}^{t,j(*)}) &\geq U_{t,j}^{\text{BS}}(\bm{\Pi}^{t,j},\bm{\alpha}^{t,j(*)}), \\
			C_{m,t,j}^{\text{user}}(\bm{\Pi}^{t,j(*)},\alpha_{m}^{t,j(*)},\bm{\alpha}^{t,j(*)}_{-m}) &\leq C_{m,t,j}^{\text{user}}(\bm{\Pi}^{t,j(*)},\alpha_m^{t,j},\bm{\alpha}^{t,j(*)}_{-m}). 
		\end{aligned} \label{SE}
	\end{equation}
\end{definition}

\begin{theorem}
	There exists an NE in Stage \Rmnum{2} and an SE for the two-stage Stackelberg game. \label{NE and SE}
\end{theorem}
\vspace{0.1cm}
\begin{proof}
	Please refer to Appendix \ref{appendix:NE and SE}.
\end{proof}

As a result of Theorem \ref{NE and SE}, we can derive the algorithmic procedure for the SE at each slot $t$ in Algorithm \ref{SE of slot}.

\section{Simulation Results}

In this section, we conduct numerical simulations to evaluate the performances of our proposed algorithms under a two-time scale framework for service cacahing, pricing and task offloading. Without loss of generality, we assume that the sizes and costs are equal for all programs $n$, i.e., $z_{n} = 50$ MB. To balance the importance of caching costs in different simulation scenarios, we define $\omega_n^j$ as 10\% of the payments received from users rather than a fixed value. Unless otherwise specified, the values of the remaining parameters are provided in Table \ref{Tab 1}.

\subsection{Task Offloading Evaluation}
We first analyze the task offloading performances of users in Stage \Rmnum{2} for the small time scale. We consider a scenario with $Z=\infty$ and assume that the BS caches all service programs. To validate the scheme of our proposed algorithm for users' cost functions, we compare the proposed threshold-based offloading (TO) strategy with the following baselines.
	\subsubsection{Complete Offloading (CO)}  
All users will offload their entire tasks to the edge server in any situation, i.e., $\alpha_m^{t,j} = 1$.
\subsubsection{Local Computing (LC)}
All users will complete their tasks through local computation in any situation, i.e., $\alpha_m^{t,j} = 0$.
\subsubsection{Random Offloading (RO)}    
Each user will randomly select a proportion $\alpha_m^{t,j} \in [0,1]$ for task offloading.

\begin{figure}[t]
	\vspace{-0.3cm}
	\centering
	\setlength{\abovecaptionskip}{-0.1cm}
	\includegraphics[width=0.4\textwidth]{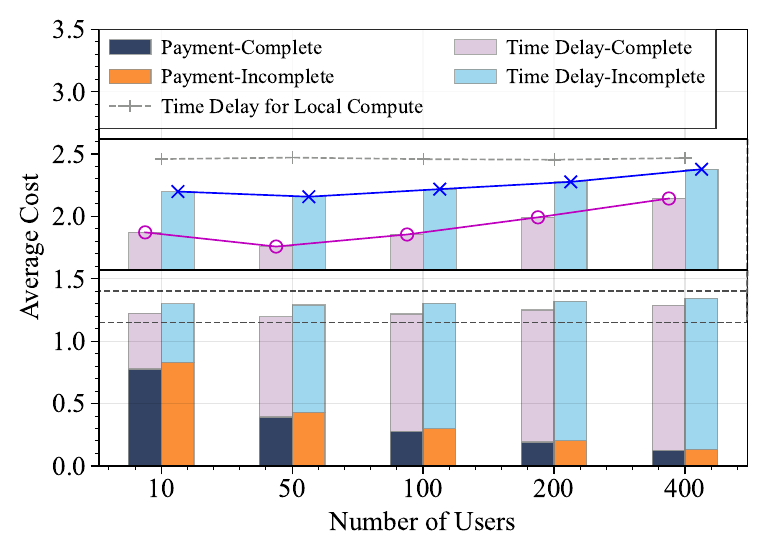}
	\caption{The average cost of users vs. the number of users  \( M \) under complete and incomplete information.}
	\label{fig:offloading2}
	\vspace{-0.2cm}
\end{figure}

In Fig. \ref{fig:offloading2}, we investigate the impact of the number of users $M$ on the average cost of all users for our proposed threshold-based offloading strategy under complete and incomplete information where $F=50$ MHz. According to the definition of the cost function (\ref{cost function of user}), the total cost for the users is composed of two parts: payment and time delay. We utilize the time delay of LC as the baseline algorithm, where its payment is consistently zero and the time delay remains unaffected by the completeness of information. We observe that an increase in $M$ leads to heightened competition for the channel rate $R_m^{t,j}$ and computation resources $F^{t,j}$. As a result of (\ref{optimal proportion}), users tend to offload fewer tasks to the edge server, thus incuring a greater time delay and lower payment. The initial downwards trend in the average cost for all users when $M$ scales up from 10 to 50 is particularly noteworthy. This trend occurs because, with an increase in $M$ based on sparse users, there is a high probability that users with a smaller characteristic parameter $\frac{\theta}{f_m}$ will enter the system. In pursuit of profit maximization, the BS tends to set lower prices according to CPTO. At this point, the decrease in the average payment outpaces the increase in the time delay. Moreover, unlike in incomplete information scenarios, users can optimize their offloading strategies with accurate knowledge of $M^{t,j}$. This approach leads to superior performance in terms of both payment and time delay under complete information.

\begin{figure}[t]
	\vspace{-0.3cm}
	\centering
	\setlength{\abovecaptionskip}{-0.1cm}
	\includegraphics[width=0.4\textwidth]{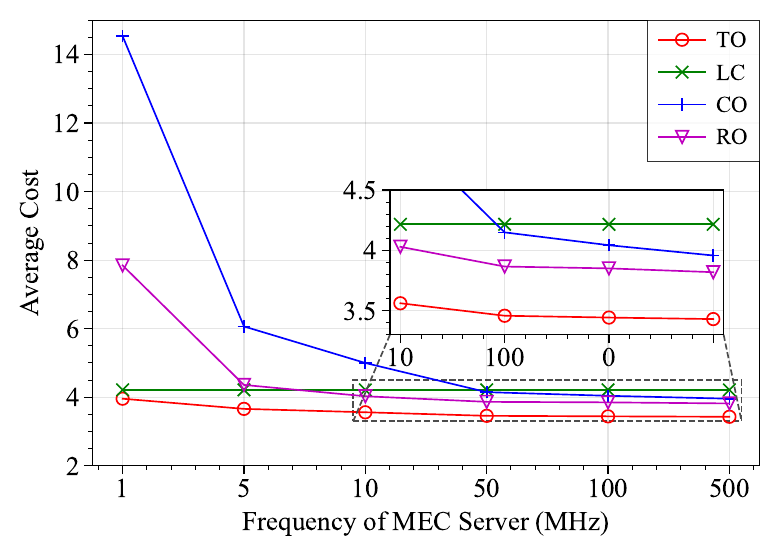}
	\caption{The average cost of users vs. the edge computation capability \(F \) for different offloading strategies.}
	\label{fig:offloading1}
	\vspace{-0.2cm}
\end{figure}

Fig. \ref{fig:offloading1} illustrates the impact of the edge computation capability $F$ on the average cost for users, where $M=50$. We observe from Fig. \ref{fig:offloading1} that our proposed TO achieves the best results. The LC curve remains unchanged because local computation does not utilize the BS's computation resources. When $F \leq 5$ MHz, the LC algorithm outperforms CO and RO while the costs of TO are 13.04\%, 39.65\% and 16.32\% lower than LC, CO and RO, respectively. As $F$ increases, the effectiveness of the three offload strategies other than LC improves accordingly. This is because users are allocated more computation resources, resulting in a decrease in the computation latency $D_{m,t,j}^\text{exe}$. Specifically, when $F = 100$ MHz, TO achieves 18.55\%, 14.84\% and 11.51\% lower costs than LC, CO and RO, repectively. In addition, when $F > 500$  MHz, the TO, CO and RO curves  exhibit a slow decline. This means that when the computing capacity of the BS is significantly greater than that of the users, further increasing $F$ has a minimal impact on the user costs and MEC system profits.

\subsection{Pricing Evaluation}
Next, we investigate the pricing strategy of the BS in Stage \Rmnum{1} for the small time scale for the case of $F = 500$ MHz. First, we show the impact of caching capacity on CPTO and SCAO. Afterwards, we investigate the impact of the number of users on five pricing algorithms---CPTO, SCAO and the following baseline algorithms---by considering two characteristics: algorithm profitability and computation complexity.
\subsubsection{PSO-Based Pricing Optimization (DPO)}
The BS utilizes the PSO algorithm to solve (\ref{utility function of program}) and obtain the optimal pricing $\bm{\Pi}^{t(*)}$ where the population size is set to $pop=100$, and the maximum number of iterations is set to $maxIter=200$ \cite{tong2023stackelberg,9861697}.
\subsubsection{Linear Pricing (LP)}
The BS determines the pricing $\bm{\Pi}^{t}$ through a linear pricing model. Specifically, the price $\pi_n^{t,j}$  is linearly correlated with the expected workload of the type-$n$ program, i.e., $\pi_n^{t,j} = \phi \mathbb{E}(r_m^{t,j}), \forall m \in \mathcal{M}_n^{t,j}$. In our simulations, we set $\phi = 3\times 10^{-9}$ \cite{ibrahim2016task,sahal2014effective}.
\subsubsection{Large Time Scale Pricing (LTSP)}
Pricing and service caching are updated on the same time scale: the BS sets the same prices as CPTO at the first time slot for each time frame, and there are no price adjustments in subsequent time slots.

\begin{figure}[t]
	\vspace{-0.3cm}
	\centering
	\setlength{\abovecaptionskip}{-0.1cm}
	\includegraphics[width=0.4\textwidth]{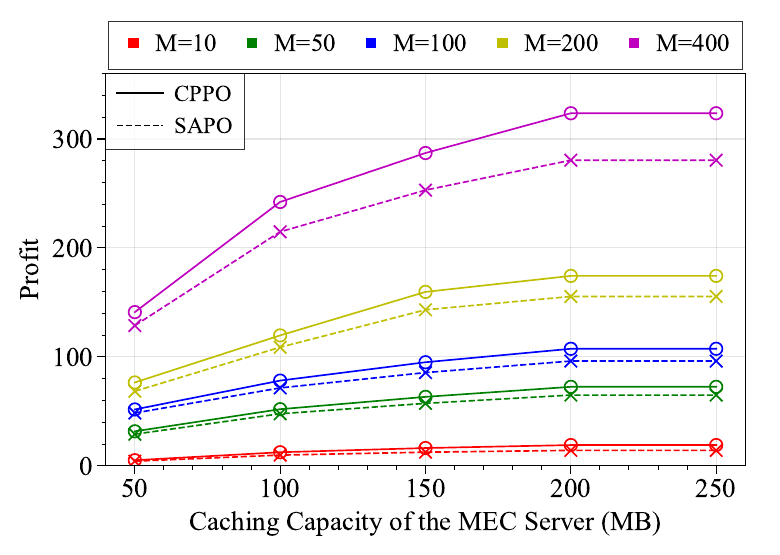}
	\caption{The profit of the BS vs. the caching capacity $Z$ for CPTO and SCAO.}
	\label{fig:Capacity}
	\vspace{-0.1cm}
\end{figure}

Fig. \ref{fig:Capacity} illustrates the impact of the caching capacity $Z$ on the performance of CPTO and SCAO algorithms in various user number scenarios. We observe that when $Z < 200 $ Mb, as $Z$ increases, both CPTO and SCAO lead to an increase in the profit of the BS. This is because  a larger caching capacity implies a more extensive range of caching choices, attracting more users to participate in task offloading. However, when the caching capacity of the BS is already large enough to cache all service programs (4 types in this case), i.e., $Z \geq 200$ Mb, further increasing $Z$  has no positive impact on the profit of the BS. When $Z$ is fixed, as the number of users increases, the BS can obtain more profits. When we further fix the number of users, the performance of CPTO consistently outperforms that of SCAO.

\begin{figure}[t]
	\vspace{-0.3cm}
	\centering
	\setlength{\abovecaptionskip}{-0.1cm}
	\includegraphics[width=0.4\textwidth]{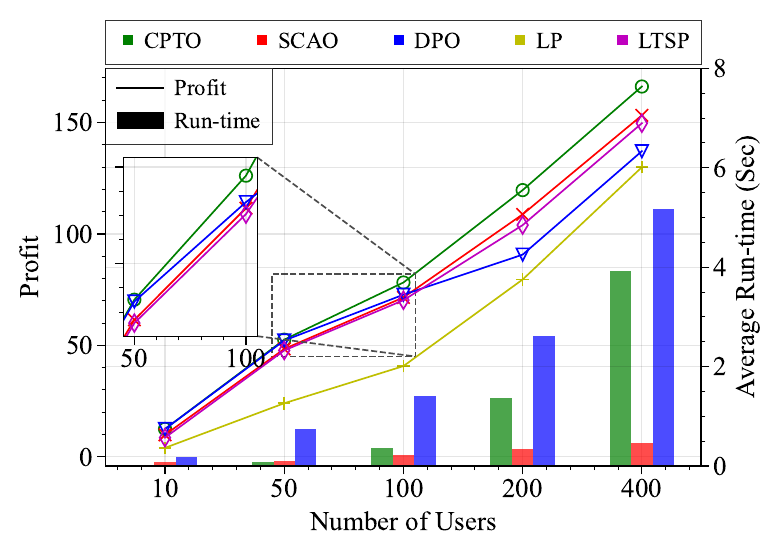}
	\caption{The profit and average run-time vs. the number of users $M$ for different pricing schemes.}
	\label{fig:DPOA_Combined}
	\vspace{-0.35cm}
	\end{figure}

In Fig. \ref{fig:DPOA_Combined}, we illustrate the impact of the number of users $M$ on algorithm profitability and computation complexity of the five pricing algorithms, where $Z=100$ Mb. We observe from Fig. \ref{fig:DPOA_Combined} that for all algorithms, when the distribution of the users' private information remains consistent, a larger number of users leads to higher profit. Our proposed CPTO algorithm consistently maintains the best performance, while the efficacy of the SCAO algorithm consistently surpasses that of the LTSP algorithm, and the profits of the LP algorithm are always the lowest. In addition, when $M < 100$, DPO yields greater profit than SCAO and LTSP, while CPTO achieves 8.61\%, 10.15\% and 0.73\% greater profits than SCAO, LTSP and DPO, respectively, when $M = 50$. However, as $M$ increases, the ascending slope of the DPO curve becomes significantly lower than those of the SCAO and LTSP curves. When $M=200$, the profit of CPTO is 10.01\%, 15.11\% and 31.94\% greater than those of SCAO, LTSP and DPO, respectively. This is because the more users participate in task offloading, the more complex the utility function of the BS becomes, making it more challenging for the PSO algorithm to optimize prices.

Furthermore, the bar graph in Fig. \ref{fig:DPOA_Combined} shows that as $M$ increases, the average run-time of the three optimization algorithms increases. In particular, DPO has a significantly higher run-time than CPTO and SCAO. The average cost of DPO is 295.01\% and 540.01\% higher than those of CPTO and SCAO, respectively, when $M=100$. We do not consider the run-time of LTSP and LP because they directly provide price schemes without the need to solve for the optimal solution of the utility function. This demonstrates that CPTO is superior to DPO in terms of both algorithm profitability and computation complexity. Additionally, SCAO is capable of obtaining satisfactory profit with the lowest run-time.

\subsection{Service Caching Evaluation}
Next, we use a real-world dataset to evaluate the proposed GNDRL-based service caching method and compare it with two baselines in terms of obtaining MEC service profit for the BS in the case of $Z=100$ Mb.
\subsubsection{Popularity-Greedy Service Caching (POSC)}
The BS greedily caches the most popular service programs based on estimated popularity information $\mathcal{Y}^{j}$ until there is not enough remaining memory to cache an additional program \cite{yan2021pricing,tao2016content}.
\subsubsection{Static Service Caching (STSC)}
The BS maintains the same cached service programs across different frames.

\begin{figure}[t]
	\centering
	\setlength{\abovecaptionskip}{-0.1cm}
	\includegraphics[width=0.4\textwidth]{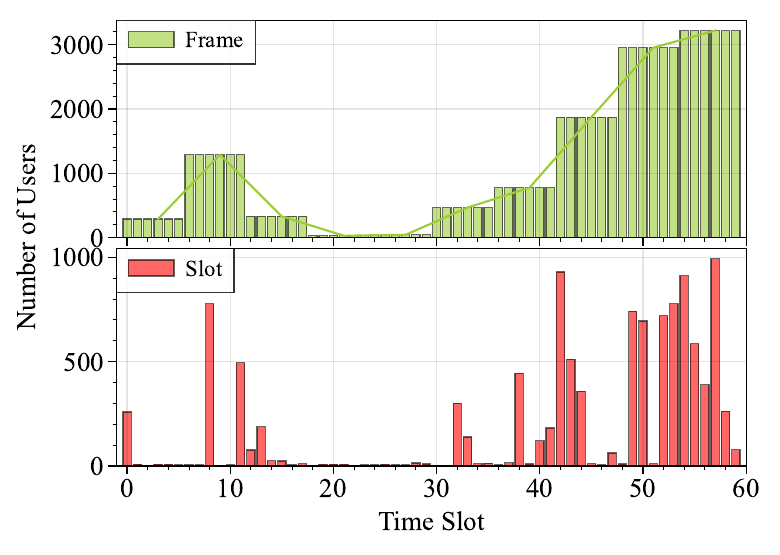}
	\caption{Distribution of the number of users over time slots and time frames.}
	\label{fig:cluster_user}
	\vspace{-0.3cm}
	\end{figure}

\begin{figure}[t]
	\vspace{0.2cm}
	\centering
	\setlength{\abovecaptionskip}{-0.1cm}
	\includegraphics[width=0.4\textwidth]{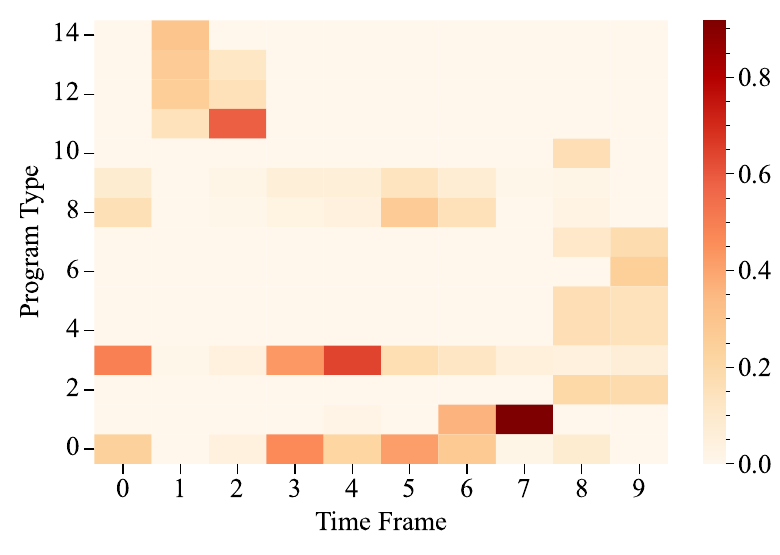}
	\caption{True popularity of 15 service programs over time frames.}
	\label{fig:cluster_heat}
	\vspace{-0.1cm}
	\end{figure}

In this simulation, we test the performances of the different caching algorithms in a real-world scenario using the ``clusterdata-2011-2" trace collected from Google clusters \cite{reiss2014google}, which represents 29 day's worth of Borg cell information starting at 19:00 EDT on Sunday May 1, 2011. We obtain the sequence of service requests from the task event table over one hour starting at 19:06 EDT on May 1. Each request is recorded as an entry, including the timestamp, task ID, job ID, number of CPU cycles and parameter size. We associate each task ID with a specific computation task, and we link the job ID of the task to a service program type \cite{ren2022adaptive,tran2019costa}. During this hour, there are a total of 60,280 request entries, with 975 different program types. The most popular program has 11,005 requests, and the least has only one request. In this simulation, we select 15 service programs with request counts ranging whithin [582, 2380]. We set the duration of a time frame to 360 s, and the duration of a time slot to 60 s; hence, we obtain 10 frames with 6 slots in each frame. We assume that each request is made by a single user, and the parameters of each user follow the distribution mentioned above. We show the changes in user number $M$ over time in Fig. \ref{fig:cluster_user} and the heat map for the true popularity $\overline{\mathcal{Y}}^j$ of the 15 service programs across each frame in Fig. \ref{fig:cluster_heat}.

\begin{figure}[t]
	\centering
	\setlength{\abovecaptionskip}{-0.1cm}
	\includegraphics[width=0.4\textwidth]{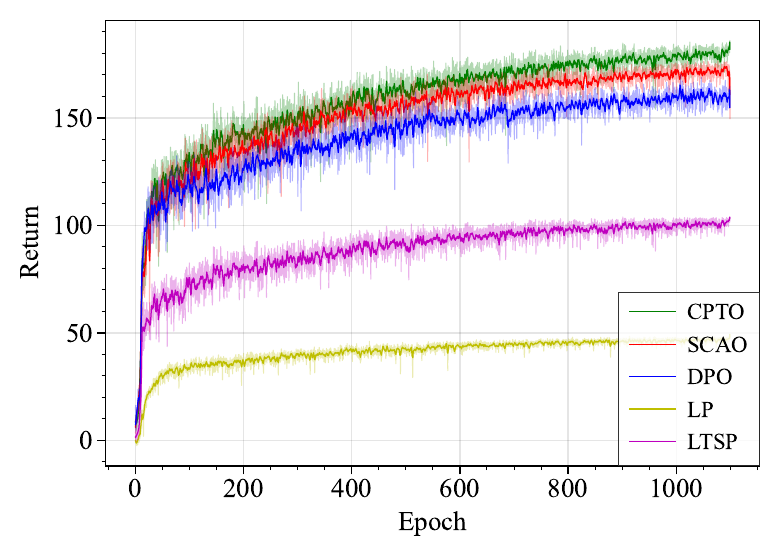}
	\caption{Convergence performance of GNDRL based on different pricing schemes.}
	\label{fig:convergence}
	\vspace{-0.4cm}
	\end{figure}

In Fig. \ref{fig:convergence}, we first evaluate the convergence performance of our proposed GNDRL based on five pricing schemes where $F = 100$ MHz. It is observed that the five curves indicate relatively fast convergence at the beginning of the training. As the iterations progress, all algorithms gradually converge to a stable level of returns. This indicates that regardless of which pricing scheme is adopted for the Stackelberg game, GNDRL can converge to a positive return, and our proposed CPTO and SCAO demonstrate superior performance in terms of returns compared to DPO, LTSP and LP. Notably, in contrast to Fig. \ref{fig:DPOA_Combined}, LTSP exhibits a markedly inferior performance compared to CPTO, SCAO and DPO in Fig. \ref{fig:convergence}. This can be attributed to the fact that, unlike the uniform distribution in prior experiments, the distribution of user characteristic parameters is more concentrated at each time slot $t$ in the real-world dataset. Consequently, the prices determined by LTSP in the initial time slot may not yield high payments in subsequent time slots, leading to lower average profits in each frame.

\begin{figure}[t]
		\centering
		\setlength{\abovecaptionskip}{-0.1cm}
		\includegraphics[width=0.4\textwidth]{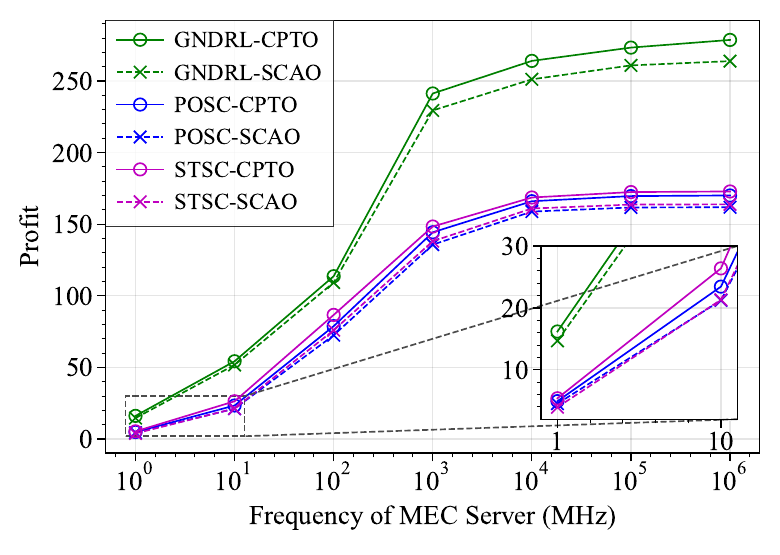}
		\caption{The profit of the BS vs. edge computation capability $F$ for different caching strategies.}
		\label{fig:cluster_caching}
		\vspace{-0.4cm}
	\end{figure}

\begin{figure*}[t]
	\vspace{-0.5cm}
	\centering
	\setlength{\abovecaptionskip}{-0.1cm}
	\subfigcapskip = -0.3cm
	\subfigure[\fontsize{6.5}{7}\selectfont GNDRL]{
		\includegraphics[width=0.31\linewidth]{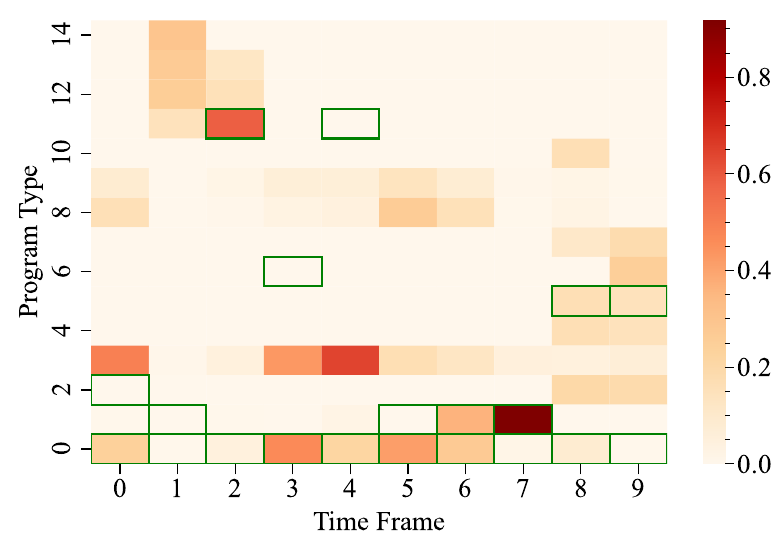}
		\label{fig:heat_DRL}
	} 
	\hfill
	\subfigure[\fontsize{6.5}{7}\selectfont POSC]{
		\includegraphics[width=0.31\linewidth]{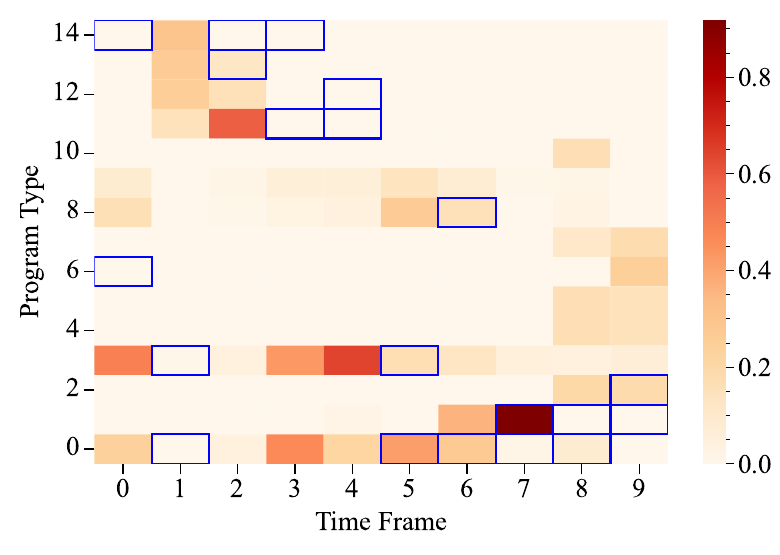}
		\label{fig:heat_Greedy}
	}
	\hfill
	\subfigure[\fontsize{6.5}{7}\selectfont STSC]{
		\includegraphics[width=0.31\linewidth]{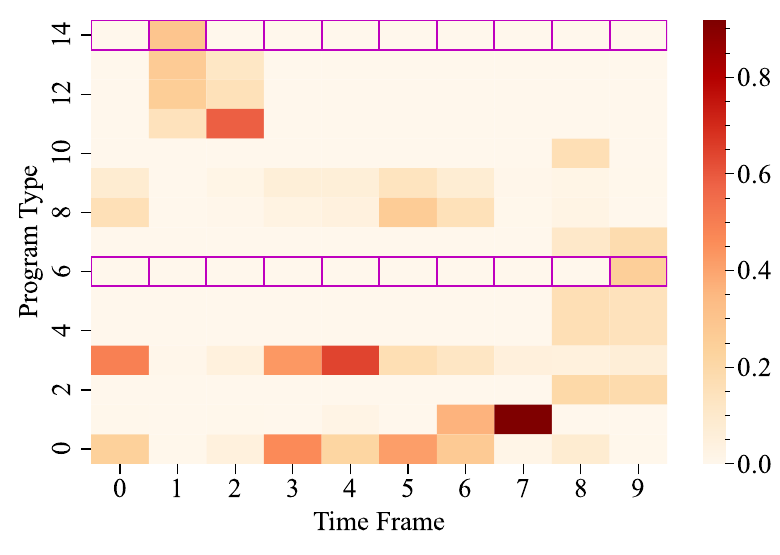}
		\label{fig:heat_Static}
	}
	\caption{Cached program results over time frames based on estimated popularity information for different caching strategies. }
	\label{fig:caching_result}
	\vspace{-0.3cm}
	
\end{figure*}

Fig. \ref{fig:cluster_caching} compares the impact of the edge computation capability $F$ on profit based on the three caching strategies. It is observed that all strategies exhibit a positive correlation across the entire range of $F$. This is because that as $F$ increases, $\delta_m^{t,j}$ in (\ref{optimal proportion}) increases, which leads to a greater proportion of offloading tasks being received by the BS; thus, the BS obtains greater profits under the same pricing scheme. When $F>10^4$ MHz, the computation resources are more than sufficient for all users, and $\delta_m^{t,j} \approx 1$. At this point, further increasing $F$ has a minimal impact on the BS's profits; hence, the growth rate of the curve significantly slows down. In addition, we observe that the strategies employing GNDRL (GNDRL-CPTO and GNDRL-SCAO) demonstrate superior profitability to the POSC and STSC strategies, where the latter two exhibit similar performance and growth trends in different $F$ scenarios. Specifically, based on CPTO, GNDRL achieves 161.67\% and 138.01\% greater profits than POSC and STSC, repectively, while based on SCAO, GNDRL achieves 169.88\% and 157.45\% greater profits than POSC and STSC. The reason is shown in Fig. \ref{fig:caching_result}(a)-(c), where we respectively display the caching results of the three algorithms with bold borders over time frames when $F=10$ MHz.

As mentioned above, the profit of the BS is composed of the total payments obtained from users and the caching costs paid to the program provider. As shown in Fig. \ref{fig:caching_result}, the caching strategy of GNDRL is mainly results in the following two behaviors: 1) GNDRL always caches the same (type-0) program over time frames to reduce the caching costs; 2) the caching results of GNDRL target the high-popularity programs with a large probability. Tis means GNDRL can obtain substantial payments from users while maintaining a low caching cost. Conversely, POSC exhibits the following two shortcomings in scenarios where program popularity changes rapidly: 1) if the most-popular programs differ between two adjacent frames, sunch as $\mathcal{Y}^j$ and $\mathcal{Y}^{j-1}$, it will lead to frequent caching adjustment, incurring significant caching costs; 2) the disparities between $\mathcal{Y}^j$ and $\overline{\mathcal{Y}}^j$ make it difficult to target the high-popularity programs in the current frame. This means that the payments yield with POSC are also low. Similar to POSC, STSC has a low target rate for high-popularity programs, but it consistently maintains the same cached programs to minimize caching costs. Therefore, the profits generated by STSC are greater than those generated by POSC.



\section{Conclusion}
In this paper, we have established a two-time scale optimization framework for dynamic popularity scenarios to jointly analyze service caching for the BS and price competition between the BS and users. For the large time scale, we have designed the GNDRL algorithm, which can achieve the optimal caching strategy based on the estimated popularity information. For the small time scale, to address price competition, we have modeled a Stackelberg game and proved the existence of the SE under incomplete information. We have provided users with the threshold-based optimal offloading strategy and designed two pricing schemes for the BS: CPTO and SCAO. The simulation results have demonstrated the superiority of the offloading and pricing strategies. Furthermore, we have utilized a real-world dataset to validate that our proposed GNDRL can generate higher profits for the BS.

\appendices


\section{The proof of Theorem \ref{NE and SE}}  \label{appendix:NE and SE}
We prove the NE and SE with the help of an EPG.
\begin{definition}[Exact Potential Game]
	A game is called an EPG if there exists a potential function $P(\cdot)$ that satisfies \cite{chen2021multi}:
	\begin{equation}
		\begin{aligned}
			U_{u,i}^{t,j}(\bm{\Pi}^{t,j},\alpha_{m}^{t,j(+)},\bm{\alpha}_{-m}^{t,j}) &- U_{u,i}^{t,j}(\bm{\Pi}^{t,j},\alpha_m^{t,j},\bm{\alpha}_{-m}^{t,j}) \\
			= P(\bm{\Pi}^{t,j},\alpha_{m}^{t,j(+)},\bm{\alpha}_{-m}^{t,j}) &- P(\bm{\Pi}^{t,j},\alpha_m^{t,j},\bm{\alpha}_{-m}^{t,j}), \\
			\forall i \in \mathcal{M},\alpha_{m}^{t,j(+)} \neq \alpha_m^{t,j}.
		\end{aligned} \label{EPG}
	\end{equation}
\end{definition}
When each user individually changes its strategy within the personal strategy space, the increase or decrease in the potential function is consistent with the corresponding user's cost function. 

The potential function reflects the total cost change of the whole game when each player alters its strategy. We map the cost functions of all users as a collective onto the potential function, and when the potential function reaches its maximum value, the EPG reaches an NE \cite{monderer1996potential,sandholm2001potential}. We define the potential function as the sum of the cost functions of all users:
\begin{equation}
	\begin{aligned}
		P(\bm{\Pi}^{t,j},\bm{\alpha}^{t,j}) &= \sum_{m=1}^M C_{m,t,j}^{\text{user}}(\bm{\Pi}^{t,j},\alpha_m^{t,j},\bm{\alpha}_{-m}^{t,j}) \\
		&= C_{m,t,j}^{\text{user}}(\bm{\Pi}^{t,j},\alpha_m^{t,j},\bm{\alpha}_{-m}^{t,j}) + \bm{C}_{-m,t,j}^{\text{user}},
	\end{aligned}
\end{equation}
where $\bm{C}_{-m,t,j}^{\text{user}}$ represents the sum of the cost functions of users other than user $m$. When user $m$ changes its offloading proportion from $\alpha_m^{t,j}$ to $\alpha_{m}^{t,j(+)}$, the change in the value of the potential function is expressed as follows:
\begin{equation}
	\begin{aligned}
		P(\bm{\Pi}^{t,j},\alpha_m^{t,j}(+),\bm{\alpha}_{-m}^{t,j}) &- P(\bm{\Pi}^{t,j},\alpha_m^{t,j},\bm{\alpha}_{-m}^{t,j}) \\
		=
		C_{m,t,j}^{\text{user}}(\bm{\Pi}^{t,j},\alpha_{m}^{t,j(+)},\bm{\alpha}_{-m}^{t,j})&- C_{m,t,j}^{\text{user}}(\bm{\Pi}^{t,j},\alpha_m^{t,j},\bm{\alpha}_{-m}^{t,j}) \\
		+ \bm{C}_{-m,t,j}^{\text{user}(+)}&-  \bm{C}_{-m,t,j}^{\text{user}}.
	\end{aligned} \label{potential function}
\end{equation}
Under the incomplete information scenario in Stage \Rmnum{2}, users determine their own offloading proportions based on the estimated offloading number (\ref{estimated number}). Therefore, the change in the offloading proportion of user $m$ does not affect the other users' utilities, i.e.,
\begin{equation}
	\begin{aligned}
		\bm{C}_{-m,t,j}^{\text{user}(+)}&-  \bm{C}_{-m,t,j}^{\text{user}} = 0.
	\end{aligned} \label{potential function 1}
\end{equation}
By substituting (\ref{potential function 1}) into (\ref{potential function}), we obtain
\begin{equation}
	\begin{aligned}
		P(\bm{\Pi}^{t,j},\alpha_{m}^{t,j(+)},\bm{\alpha}_{-m}^{t,j}) &- P(\bm{\Pi}^{t,j},\alpha_m^{t,j},\bm{\alpha}_{-m}^{t,j}) \\
		=C_{m,t,j}^{\text{user}}(\bm{\Pi}^{t,j},\alpha_{m}^{t,j(+)},\bm{\alpha}_{-m}^{t,j})&- C_{m,t,j}^{\text{user}}(\bm{\Pi}^{t,j},\alpha_m^{t,j},\bm{\alpha}_{-m}^{t,j}),
	\end{aligned}
\end{equation}
which satisfies (\ref{EPG}). This means when each user minimizes its own cost function, the overall cost for all users will not increase. Therefore, the game between users in Stage \Rmnum{2} is proven to be an EPG, and we can use the value of the potential function to represent how each user's cost function changes. As a result of (\ref{optimal proportion}), we can also conclude that $P(\bm{\Pi}^{t,j},\bm{\alpha}^{t,j})$ has an extremum value. Therefore, there exists an NE in the computation offloading game in Stage \Rmnum{2}:
\begin{equation}
	\bm{\alpha}^{t,j(*)} (\bm{\Pi}^{t,j})=\argmin_{\bm{\alpha}^{t,j}} \, C_{m,t,j}^{\text{user}}(\bm{\Pi}^{t,j},\bm{\alpha}^{t,j}).
\end{equation}

On the basis of $\bm{\alpha}^{t,j(*)}(\bm{\Pi}^{t,j})$, we already proved the existence of the maximum value of the BS's utility function (\ref{utility function of BS}) in slot $t$ in subsection \ref{section:pricing}. In addition, (\ref{utility function of BS}) can be regarded as the potential function of a single-player game that is continuous and has an extremum, and the NE of this game is
\begin{equation}
	\bm{\Pi^}{t,j(*)}=\argmax_{\bm{\Pi}^{t,j}} \, U_{t,j}^{\text{BS}}(\bm{\Pi}^{t,j},\bm{\alpha}^{t,j(*)} (\bm{\Pi}^{t,j})).
\end{equation}

According to (\ref{SE}), ($\bm{\Pi}^{t,j(*)},\bm{\alpha}^{t,j(*)}$) constitutes the SE of the Stackelberg game for the small time scale. Theorem \ref{NE and SE} is proved.

\normalem	
\bibliographystyle{IEEEtran} 
\bibliography{references}

\begin{thebibliography}{10}
\providecommand{\url}[1]{#1}
\csname url@samestyle\endcsname
\providecommand{\newblock}{\relax}
\providecommand{\bibinfo}[2]{#2}
\providecommand{\BIBentrySTDinterwordspacing}{\spaceskip=0pt\relax}
\providecommand{\BIBentryALTinterwordstretchfactor}{4}
\providecommand{\BIBentryALTinterwordspacing}{\spaceskip=\fontdimen2\font plus
\BIBentryALTinterwordstretchfactor\fontdimen3\font minus \fontdimen4\font\relax}
\providecommand{\BIBforeignlanguage}[2]{{%
\expandafter\ifx\csname l@#1\endcsname\relax
\typeout{** WARNING: IEEEtran.bst: No hyphenation pattern has been}%
\typeout{** loaded for the language `#1'. Using the pattern for}%
\typeout{** the default language instead.}%
\else
\language=\csname l@#1\endcsname
\fi
#2}}
\providecommand{\BIBdecl}{\relax}
\BIBdecl

\bibitem{10261251}
Y.~Fang, M.~Li, F.~R. Yu, P.~Si, R.~Yang, C.~Gao, and Y.~Sun, ``Parallel offloading and resource optimization for multi-hop ad hoc network-enabled cbtc with mobile edge computing,'' \emph{IEEE Transactions on Vehicular Technology}, vol.~73, no.~2, pp. 2684--2698, 2024.

\bibitem{sun2020bandwidth}
Y.~Sun, Z.~Chen, M.~Tao, and H.~Liu, ``Bandwidth gain from mobile edge computing and caching in wireless multicast systems,'' \emph{IEEE Transactions on Wireless Communications}, vol.~19, no.~6, pp. 3992--4007, 2020.

\bibitem{el2019joint}
E.~El~Haber, T.~M. Nguyen, and C.~Assi, ``Joint optimization of computational cost and devices energy for task offloading in multi-tier edge-clouds,'' \emph{IEEE Transactions on Communications}, vol.~67, no.~5, pp. 3407--3421, 2019.

\bibitem{IDC2024}
M.~Torchia and M.~Shirer, ``New idc spending guide forecasts edge computing investments will reach \$232 billion in 2024,'' IDC Media Center, 2024, available online: \url{https://www.idc.com/getdoc.jsp?containerId=prUS51960324}.

\bibitem{yan2021pricing}
J.~Yan, S.~Bi, L.~Duan, and Y.-J.~A. Zhang, ``Pricing-driven service caching and task offloading in mobile edge computing,'' \emph{IEEE Transactions on Wireless Communications}, vol.~20, no.~7, pp. 4495--4512, 2021.

\bibitem{zhao2018red}
T.~Zhao, I.-H. Hou, S.~Wang \emph{et~al.}, ``Red/led: An asymptotically optimal and scalable online algorithm for service caching at the edge,'' \emph{IEEE Journal on Selected Areas in Communications}, vol.~36, no.~8, pp. 1857--1870, 2018.

\bibitem{yan2019optimal}
J.~Yan, S.~Bi, Y.~J. Zhang, and M.~Tao, ``Optimal task offloading and resource allocation in mobile-edge computing with inter-user task dependency,'' \emph{IEEE Transactions on Wireless Communications}, vol.~19, no.~1, pp. 235--250, 2019.

\bibitem{tong2023stackelberg}
Z.~Tong, X.~Deng, J.~Mei \emph{et~al.}, ``Stackelberg game-based task offloading and pricing with computing capacity constraint in mobile edge computing,'' \emph{Journal of Systems Architecture}, vol. 137, p. 102847, 2023.

\bibitem{thai2019workload}
M.~T. Thai, Y.-D. Lin, Y.-C. Lai \emph{et~al.}, ``Workload and capacity optimization for cloud-edge computing systems with vertical and horizontal offloading,'' \emph{IEEE Transactions on Network and Service Management}, vol.~17, no.~1, pp. 227--238, 2019.

\bibitem{fudenberg1991game}
D.~Fudenberg and J.~Tirole, \emph{Game theory}.\hskip 1em plus 0.5em minus 0.4em\relax MIT Press, 1991.

\bibitem{wang2023delay}
D.~Wang, W.~Wang, H.~Gao, Z.~Zhang, and Z.~Han, ``Delay-optimal computation offloading in large-scale multi-access edge computing using mean field game,'' \emph{IEEE Transactions on Wireless Communications}, 2023.

\bibitem{yeganeh2023novel}
S.~Yeganeh, A.~B. Sangar, and S.~Azizi, ``A novel q-learning-based hybrid algorithm for the optimal offloading and scheduling in mobile edge computing environments,'' \emph{Journal of Network and Computer Applications}, vol. 214, p. 103617, 2023.

\bibitem{fan2022joint}
W.~Fan, J.~Han, Y.~Su \emph{et~al.}, ``Joint task offloading and service caching for multi-access edge computing in wifi-cellular heterogeneous networks,'' \emph{IEEE Transactions on Wireless Communications}, vol.~21, no.~11, pp. 9653--9667, 2022.

\bibitem{liu2023dependent}
S.~Liu, Y.~Yu, X.~Lian, Y.~Feng, C.~She, P.~L. Yeoh, L.~Guo, B.~Vucetic, and Y.~Li, ``Dependent task scheduling and offloading for minimizing deadline violation ratio in mobile edge computing networks,'' \emph{IEEE Journal on Selected Areas in Communications}, vol.~41, no.~2, pp. 538--554, 2023.

\bibitem{dong2023joint}
C.~Dong, Y.~Tian, Z.~Zhou, W.~Wen, and X.~Chen, ``Joint power allocation and task offloading for reliability-aware services in noma-enabled mec,'' \emph{IEEE Transactions on Wireless Communications}, 2023.

\bibitem{wang2019profit}
Q.~Wang, S.~Guo, J.~Liu \emph{et~al.}, ``Profit maximization incentive mechanism for resource providers in mobile edge computing,'' \emph{IEEE Transactions on Services Computing}, vol.~15, no.~1, pp. 138--149, 2019.

\bibitem{liwang2022unifying}
M.~Liwang, R.~Chen, X.~Wang, and X.~Shen, ``Unifying futures and spot market: Overbooking-enabled resource trading in mobile edge networks,'' \emph{IEEE Transactions on Wireless Communications}, vol.~21, no.~7, pp. 5467--5485, 2022.

\bibitem{tong2024stackelberg}
Z.~Tong, Y.~Zhang, J.~Mei, W.~Ai, K.~Li, and K.~Li, ``Stackelberg game-based bandwidth allocation and resource pricing for multi-user in mec system,'' \emph{IEEE Internet of Things Journal}, 2024.

\bibitem{chen2022price}
Z.~Chen, Q.~Ma, L.~Gao, and X.~Chen, ``Price competition in multi-server edge computing networks under saa and siq models,'' \emph{IEEE Transactions on Mobile Computing}, 2022.

\bibitem{huang2023pricing}
X.~Huang, T.~Huang, W.~Zhang, C.~K. Yeo, S.~Zhao, and G.~Zhang, ``Pricing optimization in mec systems: Maximizing resource utilization through joint server configuration and dynamic operation,'' \emph{IEEE Transactions on Mobile Computing}, 2023.

\bibitem{chu2023joint}
W.~Chu, X.~Jia, Z.~Yu \emph{et~al.}, ``Joint service caching, resource allocation and task offloading for mec-based networks: A multi-layer optimization approach,'' \emph{IEEE Transactions on Mobile Computing}, 2023.

\bibitem{qin2023joint}
L.~Qin, H.~Lu, Y.~Lu, C.~Zhang, and F.~Wu, ``Joint optimization of base station clustering and service caching in user-centric mec,'' \emph{IEEE Transactions on Mobile Computing}, 2023.

\bibitem{xu2022stable}
Z.~Xu, Q.~Xia, L.~Wang, P.~Zhou, J.~C.~S. Lui, W.~Liang, W.~Xu, and G.~Wu, ``Stable service caching in mecs of hierarchical service markets with uncertain request rates,'' \emph{IEEE Transactions on Mobile Computing}, vol.~22, no.~7, pp. 4279--4296, 2023.

\bibitem{farhadi2021service}
V.~Farhadi, F.~Mehmeti, T.~He \emph{et~al.}, ``Service placement and request scheduling for data-intensive applications in edge clouds,'' \emph{IEEE/ACM Transactions on Networking}, vol.~29, no.~2, pp. 779--792, 2021.

\bibitem{ren2022adaptive}
D.~Ren, X.~Gui, and K.~Zhang, ``Adaptive request scheduling and service caching for mec-assisted iot networks: An online learning approach,'' \emph{IEEE Internet of Things Journal}, vol.~9, no.~18, pp. 17\,372--17\,386, 2022.

\bibitem{yao2023cooperative}
Z.~Yao, S.~Xia, Y.~Li, and G.~Wu, ``Cooperative task offloading and service caching for digital twin edge networks: A graph attention multi-agent reinforcement learning approach,'' \emph{IEEE Journal on Selected Areas in Communications}, 2023.

\bibitem{diamanti2022incentive}
M.~Diamanti, P.~Charatsaris, E.~E. Tsiropoulou, and S.~Papavassiliou, ``Incentive mechanism and resource allocation for edge-fog networks driven by multi-dimensional contract and game theories,'' \emph{IEEE Open Journal of the Communications Society}, vol.~3, pp. 435--452, 2022.

\bibitem{gu2020deep}
B.~Gu, X.~Zhang, Z.~Lin, and M.~Alazab, ``Deep multiagent reinforcement-learning-based resource allocation for internet of controllable things,'' \emph{IEEE Internet of Things Journal}, vol.~8, no.~5, pp. 3066--3074, 2020.

\bibitem{poularakis2019joint}
K.~Poularakis, J.~Llorca, A.~M. Tulino, I.~Taylor, and L.~Tassiulas, ``Joint service placement and request routing in multi-cell mobile edge computing networks,'' in \emph{IEEE INFOCOM 2019 - IEEE Conference on Computer Communications}, 2019, pp. 10--18.

\bibitem{9861697}
Z.~Xiao, J.~Shu, H.~Jiang, J.~C.~S. Lui, G.~Min, J.~Liu, and S.~Dustdar, ``Multi-objective parallel task offloading and content caching in d2d-aided mec networks,'' \emph{IEEE Transactions on Mobile Computing}, vol.~22, no.~11, pp. 6599--6615, 2023.

\bibitem{ibrahim2016task}
E.~Ibrahim, N.~A. El-Bahnasawy, and F.~A. Omara, ``Task scheduling algorithm in cloud computing environment based on cloud pricing models,'' in \emph{2016 World Symposium on Computer Applications \& Research (WSCAR)}, 2016, pp. 65--71.

\bibitem{sahal2014effective}
R.~Sahal and F.~A. Omara, ``Effective virtual machine configuration for cloud environment,'' in \emph{2014 9th International Conference on Informatics and Systems}.\hskip 1em plus 0.5em minus 0.4em\relax IEEE, 2014, pp. PDC--15.

\bibitem{tao2016content}
M.~Tao, E.~Chen, H.~Zhou, and W.~Yu, ``Content-centric sparse multicast beamforming for cache-enabled cloud ran,'' \emph{IEEE Transactions on Wireless Communications}, vol.~15, no.~9, pp. 6118--6131, 2016.

\bibitem{reiss2014google}
C.~Reiss, J.~Wilkes, and J.~Hellerstein, ``Google cluster-usage traces: Format+ schema version 2.1,'' \emph{Google, Mountain View, CA, USA, Tech. Rep}, 2011, [Online]. Available: \url{https://github.com/google/cluster-data}.

\bibitem{tran2019costa}
T.~X. Tran, K.~Chan, and D.~Pompili, ``Costa: Cost-aware service caching and task offloading assignment in mobile-edge computing,'' in \emph{2019 16th annual IEEE international conference on sensing, communication, and networking (SECON)}.\hskip 1em plus 0.5em minus 0.4em\relax IEEE, 2019, pp. 1--9.

\bibitem{chen2021multi}
J.~Chen, Q.~Wu, Y.~Xu, N.~Qi, T.~Fang, L.~Jia, and C.~Dong, ``A multi-leader multi-follower stackelberg game for coalition-based uav mec networks,'' \emph{IEEE Wireless Communications Letters}, vol.~10, no.~11, pp. 2350--2354, 2021.

\bibitem{monderer1996potential}
D.~Monderer and L.~S. Shapley, ``Potential games,'' \emph{Games and economic behavior}, vol.~14, no.~1, pp. 124--143, 1996.

\bibitem{sandholm2001potential}
W.~H. Sandholm, ``Potential games with continuous player sets,'' \emph{Journal of Economic theory}, vol.~97, no.~1, pp. 81--108, 2001.

\end{thebibliography}

\end{document}